\documentclass{article}

\usepackage{microtype}
\usepackage{graphicx}
\usepackage{subfigure}
\usepackage{booktabs} 
\usepackage{caption}
\usepackage{enumitem}
\usepackage{makecell}
\usepackage{graphicx}
\usepackage{tikz}
\usepackage{pgfplots}
\usepackage{amsfonts}
\usepackage{hyperref}

\usepackage{hyperref}

\usepackage[accepted]{icml2020}

\icmltitlerunning{DropNet: Reducing Neural Network Complexity via Iterative Pruning}

\begin{document}

\twocolumn[
\icmltitle{DropNet: Reducing Neural Network Complexity via Iterative Pruning}



\begin{icmlauthorlist}
\icmlauthor{John Tan Chong Min}{nus}
\icmlauthor{Mehul Motani}{nus}
\end{icmlauthorlist}

\icmlaffiliation{nus}{Department of Electrical and Computer Engineering, National University of Singapore}

\icmlcorrespondingauthor{John Tan Chong Min}{johntancm@u.nus.edu}
\icmlcorrespondingauthor{Mehul Motani}{motani@u.nus.edu}

\icmlkeywords{Machine Learning, Pruning, Iterative Pruning, ICML}

\vskip 0.3in
]

\printAffiliationsAndNotice{} 

\begin{abstract}
Modern deep neural networks require a significant amount of computing time and power to train and deploy, which limits their usage on edge devices. Inspired by the iterative weight pruning in the Lottery Ticket Hypothesis \cite{Frankle2018}, we propose \textit{DropNet}, an iterative pruning method which prunes nodes/filters to reduce network complexity. \textit{DropNet} iteratively removes nodes/filters with the lowest average post-activation value across all training samples. Empirically, we show that \textit{DropNet} is robust across diverse scenarios, including MLPs and CNNs using the MNIST, CIFAR-10 and Tiny ImageNet datasets. We show that up to 90\% of the nodes/filters can be removed without any significant loss of accuracy. The final pruned network performs well even with reinitialization of the weights and biases. \textit{DropNet} also has similar accuracy to an oracle which greedily removes nodes/filters one at a time to minimise training loss, highlighting its effectiveness. 

\end{abstract}

\section{Introduction}
The surprising effectiveness of neural networks in domains such as image recognition in ImageNet \cite{ILSVRC15} has inspired much recent research. Current state-of-the-art deep models include Transformers \cite{DBLP:journals/corr/VaswaniSPUJGKP17} for language modelling, InceptionNet \cite{DBLP:journals/corr/SzegedyLJSRAEVR14} for image modelling, and ResNets \cite{DBLP:journals/corr/HeZRS15} which include over 100 layers. In fact, the number of configurable parameters per model has risen significantly from hundreds to tens of millions. The increased computational complexity required for modern neural networks has made deployment in edge devices challenging. 

\textbf{Previous Work on Complexity Reduction:} Current methods of reducing complexity include quantization to 16-bit \cite{gupta2015deep}, group L1 or L2 regularization \cite{Alemu2019}, node pruning \cite{castellano1997, Zhang2010, Alvarez2016, wen2016learning}, filter pruning for CNNs \cite{DBLP:journals/corr/LiKDSG16, wen2016learning, Alvarez2016, he2017channel, liu2017learning}, weight pruning using magnitude-based methods \cite{han2015learning} or second-order Hessian-based methods such as Optimal Brain Damage \cite{NIPS1989_250} or Optimal Brain Surgeon \cite{NIPS1992_647}. 

\textbf{Previous Work on Node/Filter Pruning:}
For node pruning, previous work includes introducing regularization terms based on input weights using group lasso to the loss function \cite{Alvarez2016, wen2016learning} and selecting the least important node to drop based on mutual information \cite{Zhang2010}. In CNNs, previous work includes layer-wise pruning such as layer-wise lasso regression on filters \cite{he2017channel}, and global pruning methods such as pruning filters with the lowest sum of absolute input weights globally \cite{DBLP:journals/corr/LiKDSG16}, using second-order Taylor expansion to prune unimportant filters \cite{molchanov2016pruning, lin2018accelerating}, introducing structured sparsity using a particle filter approach \cite{anwar2017structured}, or to repeatedly perform an L1 regularization of the batch normalization layer's scaling factor in CNNs \cite{liu2017learning}.

\textbf{Previous Work on Iterative Pruning:} Most techniques that can perform one-shot pruning can also be applied recursively using iterative pruning. It has been shown that iterative pruning achieves better performance than one-shot pruning \cite{DBLP:journals/corr/LiKDSG16}. More notably, iterative pruning with reinitialization can reduce parameter counts by over 90\% \cite{Frankle2018}. Such iterative pruning methods shed new insight into how more effective pruning approaches might be developed.

\textbf{Comparison with Similar Work:} \textit{DropNet} iteratively removes nodes/filters with the lowest average post-activation values across all training samples. Similar work to ours includes removing nodes with the highest average percentage of zero activation values across a validation set - Average Percentage of Zeros (APoZ) \cite{hu2016data}, which measures sparsity of a node's activations. In contrast, \textit{DropNet} (i) utilizes average magnitude of post-activation values, and (ii) does so over the training set. Another similar metric is to prune channels to a filter using variance of post-activation values \cite{polyak2015channel}. \textit{DropNet} instead prunes the entire filter using average post-activation values.

\textbf{Comparison with Dropout:} Dropout \cite{dropout} randomly drops a fraction $p$ of nodes during training, but keeps the entire network intact during test time. \textit{DropNet}, however, drops nodes/filters permanently during training time and test time.

\textbf{Our Contributions:}
\begin{itemize}[leftmargin=*,nosep]
\item We propose \textit{DropNet}, an iterative node/filter pruning approach with reinitialization, which iteratively removes nodes/filters with the lowest average post-activation value across all training samples (either layer-wise or globally) and, hence, reduces network complexity. To the best of our knowledge, our method is the first to prune both MLPs and CNNs based on the average post-activation values of nodes/filters, which utilizes both the information about the input weights as well as the input data to make an informed selection of the nodes/filters to drop.

\item \textit{DropNet} achieves a robust performance across a wide range of scenarios compared to several benchmark metrics.  \textit{DropNet} achieves similar performance to an {\em oracle} which greedily removes nodes/filters one at a time to minimise training loss.

\item \textit{DropNet} does not require special initialization of weights and biases (unlike \cite{Frankle2018}). It is shown in subsequent experiments that a random initialization of the pruned model will do just as well as original initialization. This means the architecture pruned by \textit{DropNet} can be readily deployed using off-the-shelf machine learning libraries and hardware.

\end{itemize}

\section{{\em DropNet} Algorithm}

In this section, we describe the \textit{DropNet} algorithm and discuss its properties. 

Similar to the Lottery Ticket Hypothesis \cite{Frankle2018}, which iteratively drops weights with reinitialization, \textit{DropNet} applies iterative dropping for nodes/filters with reinitialization.

{\bf Model:} Consider a dense feed-forward neural network $f(x; n)$ with initial configuration of weights and biases $\theta = \theta_0$ and initial configuration of nodes/filters $n$. Let $f$ reach minimum validation loss $l$ with test accuracy $a$ when optimizing with stochastic gradient descent (SGD) on a training set. Consider also training $f(x; m\odot n)$ with a mask $m \in \{0,1\}^{|n|}$ on its nodes/filters such that its initialization is $f(x; m\odot n)$, where $m\odot n$ is an element-wise multiplication between $m$ and $n$. Let $f$ with the mask reach minimum validation loss $l'$ with with test accuracy $a'$. 

{\bf Problem:} Find a subnetwork $f(x; m\odot n)$ such that $a' \approx a$ \textit{(similar accuracy)} and $||m||_0 \ll n$ \textit{(fewer parameters)}.

\begin{figure}[!t]
    \includegraphics[width=\linewidth]{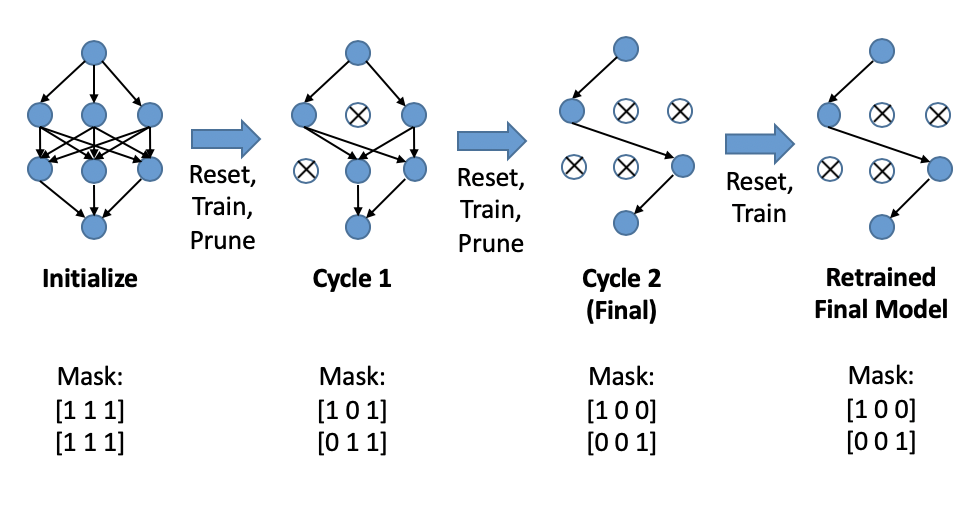}
    \caption{Illustration of DropNet algorithm as described in Algorithm \ref{alg:DropNet}. The mask is initially set to all 1s, meaning all nodes/filters are present. As the training cycle progresses, more and more nodes/filters are dropped. The final model at cycle 2 is then reset to initial state and retrained to give the final parameters.}
    \label{fig:DropNet_Algo}
\end{figure}

{\bf Algorithm:} The proposed iterative pruning algorithm is shown in Algorithm~\ref{alg:DropNet}. \textit{DropNet} applies Algorithm ~\ref{alg:DropNet} with the following metric: the lowest average post-activation value across all training samples (either layer-wise or globally). An example of the training cycle using Algorithm ~\ref{alg:DropNet} is shown in Fig. \ref{fig:DropNet_Algo}.

\begin{algorithm}[!t]
	\caption{Iterative Pruning Algorithm}
	\label{alg:DropNet}
	\begin{algorithmic}
		\STATE {\bfseries Input:} Neural network with initial state $\theta_0$, inital nodes/filters $n$, pruning metric. 
		\STATE {\bfseries Hyperparameters:} Training iterations $j$, pruning fraction $p \in (0,1]$, maximum loss factor $\kappa \in (0,1]$
	    \STATE Initialize starting mask $m = \{1\}^{|n|}$
		\REPEAT
		\STATE 1. Revert network to initial state $\theta_0$
		\STATE 2. Apply mask to nodes/filters: $f(x; m\odot n)$
		\STATE 3. Train network for at most $j$ iterations until early stopping
		\STATE 4. Apply pruning metric to choose a fraction $p$ of nodes/filters to drop and update $m$

		\UNTIL{validation accuracy $a' \leq \kappa a$}
		\STATE Run steps 1 to 3
	\end{algorithmic}
\end{algorithm}

{\bf Expected Absolute Value of a Node:} Unlike weights, which can be of the same value for all training samples, nodes will change their post-activation values according to the input training samples. Hence, we use a node's \textit{expected absolute value} across all training samples ($x_1, x_2, ..., x_t$) to evaluate its importance. For each node $a_i, i \in \mathbb{Z^+}, 1 \leq i \leq n$, out of all $n$ nodes in the network, we have its \textit{expected absolute value} as:
$$E(a_i) = \frac{1}{t}\sum_{j=1}^{t}|V(a_i|x_j)|,$$
where $V(a_i|x_j)$ is the post-activation value of the node $a_i$ with input sequence $x_j$.

{\bf Expected Absolute Value of a Filter:} For CNNs, the filters are comprised of a set of constituent nodes. Choosing which filter to drop is thus equivalent to choosing a set of constituent nodes to drop. Hence, in order to evaluate the importance of each filter $f_i, i \in \mathbb{Z^+}, 1 \leq i \leq n$, out of all $n$ filters in the network, we take the average of the \textit{expected absolute value} of all its constituent nodes $a_1, a_2, ..., a_r$, $r \in \mathbb{Z}^+$. That is:
$$E(f_i) = \frac{1}{r}\sum_{j=1}^{r}|E(a_j)|,$$
where $E(f_i)$ is the \textit{expected absolute value} of the filter $f_i$ across all training samples $x_1, x_2, ..., x_t$.

\textbf{Intuition:} We propose two reasons for the competitiveness of dropping nodes/filters with lowest average post-activation value across all training samples. (i) Firstly, Rectified Linear Unit ($ReLU$) activation will lead to inactive nodes which do not ``fire" once the value of node reaches 0 or below. Hence, if a node does not fire most of the time, it will have a low expected absolute value and removing it will affect only a small amount of classifications when its post-activation value is non-zero. (ii) Secondly, during backpropagation, the input weights of a node with low expected absolute value will be updated by only a small amount, which means that the node is less adaptive to learning from the inputs. Removing these less adaptive nodes should impact classification accuracy less than removing more adaptive ones. 

\section{Methodology}

In order to evaluate the effectiveness of the \textit{DropNet} algorithm, we test it empirically using MLPs and CNNs on MNIST \cite{lecun2010mnist},  CIFAR-10 \cite{Krizhevsky09learningmultiple} and Tiny ImageNet (taken from https://tiny-imagenet.herokuapp.com, results in Supplementary Material) datasets. 

\subsection{Pruning Metrics}

The pruning metrics we consider are listed in Table \ref{table:metrics}. At the end of each training cycle in Algorithm \ref{alg:DropNet}, we use the metric to evaluate the importance score for each node/filter, and drop the nodes/filters with the lowest importance scores. Note that ties are broken randomly.

\begin{table}[t]
	\caption{Pruning Metrics}
	\label{table:metrics}
	\vskip 0.15in
	\begin{center}
		\begin{small}
			\begin{sc}
				\begin{tabular}{ccc}
					\toprule
					Metric & Importance Score & Type\\
					\midrule
					minimum   & $E(a_i)$ or $E(f_i)$ & Global\\
					maximum  & $-E(a_i)$ or $-E(f_i)$ & Global\\
					random & 0 & Global\\
					minimum\_layer   & $E(a_i)$ or $E(f_i)$ & Layer-wise\\
					maximum\_layer  & $-E(a_i)$ or $-E(f_i)$ & Layer-wise\\
					random\_layer & 0 & Layer-wise\\
					\bottomrule
				\end{tabular}
			\end{sc}
		\end{small}
	\end{center}
	\vskip -0.1in
\end{table}

The \texttt{minimum}, \texttt{maximum} and \texttt{random} metrics prune a fraction $p$ of nodes/filters globally. The \texttt{minimum} metric drops a fraction $p$ the nodes/filters globally with the lowest post-activation values. The \texttt{maximum} metric serves as a comparison to the \texttt{minimum} metric to compare the effectiveness of the metric. The \texttt{random} metric, which prunes a fraction $p$ of nodes randomly, serves as a control. 

We consider layer-wise pruning metrics. These metrics are termed \texttt{minimum\_layer}, \texttt{maximum\_layer} and \texttt{random\_layer}, and they prune a fraction $p$ of nodes/filters layer-wise. 

\textit{DropNet} utilizes Algorithm \ref{alg:DropNet} with either the \texttt{minimum} or \texttt{minimum\_layer} metrics. As will be shown in our experimental results, these metrics prove to be quite competitive for different scenarios.

\subsection{Experiment Details}

\textbf{Train-Validation-Test Split:} For MNIST, the dataset is split into 54000 training, 6000 validation and 10000 testing samples. For CIFAR-10, the dataset is split into 45000 training, 5000 validation and 10000 testing samples.

\textbf{Pre-processing:} The input pixel values are scaled to be between 0 and 1. 

\textbf{Activation Function:} The model activation functions are all \textit{ReLU}, except the final classification layer where it is softmax in order to choose one out of multiple classes.

\textbf{Optimization Function:} The optimization function used is SGD with a learning rate of 0.1.

\textbf{Loss Function:} The loss function used is cross entropy.

\textbf{Training Runs:} The experiments are repeated over 15 runs, each with a different initial random seed. To serve as comparison between the various metrics, the average accuracy against the fraction of nodes/filters remaining across all 15 runs are plotted, together with the error bars denoting the 95\% confidence interval. 

\textbf{Training Cycles:} The masks are applied at the start of each training cycle, which comprises 100 epochs, with early stopping using validation loss with patience of 5 epochs. Over each training cycle, a fraction $p = 0.2$ of the nodes are dropped.

\begin{figure*}[!t]
\begin{center}
\hfill
\begin{minipage}[t]{0.2\textwidth}
\centering
    \includegraphics[height=3in]{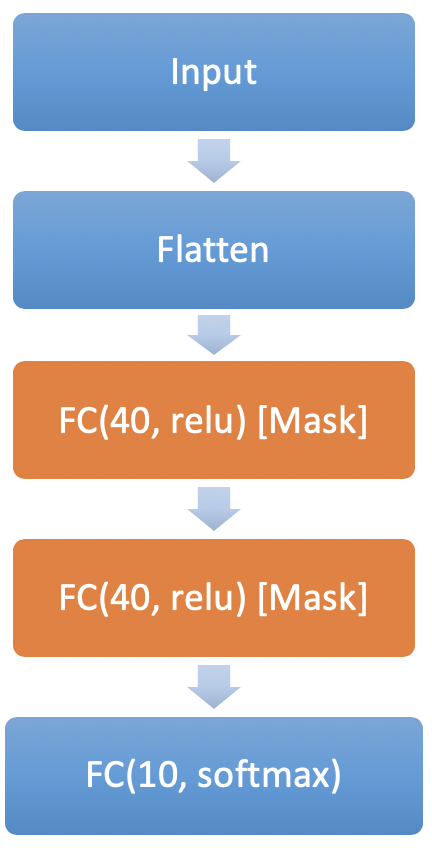}
    \centerline{Model A}
\end{minipage}
\hfill
\begin{minipage}[t]{0.35\textwidth}
\centering
    \includegraphics[height=3in]{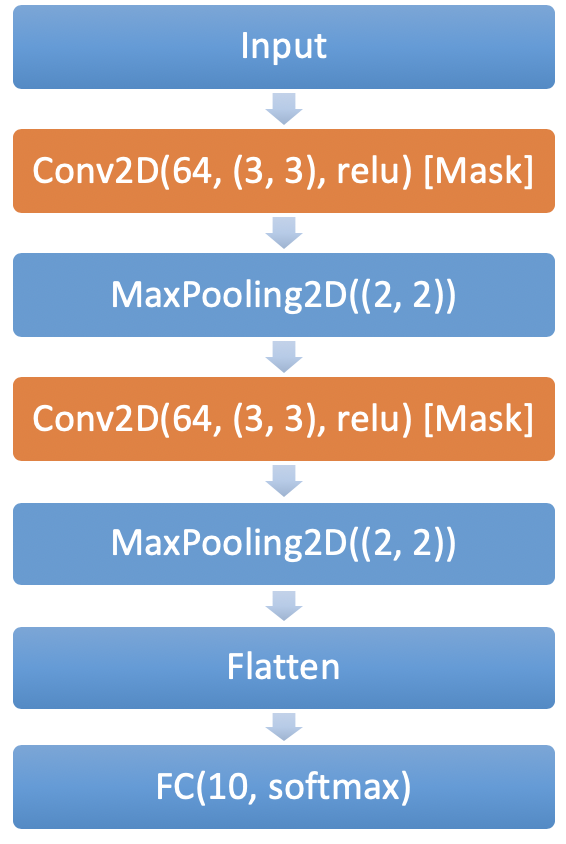}
    \centerline{Model B}
\end{minipage}
\begin{minipage}[t]{0.35\textwidth}
\centering
    \includegraphics[height=3in]{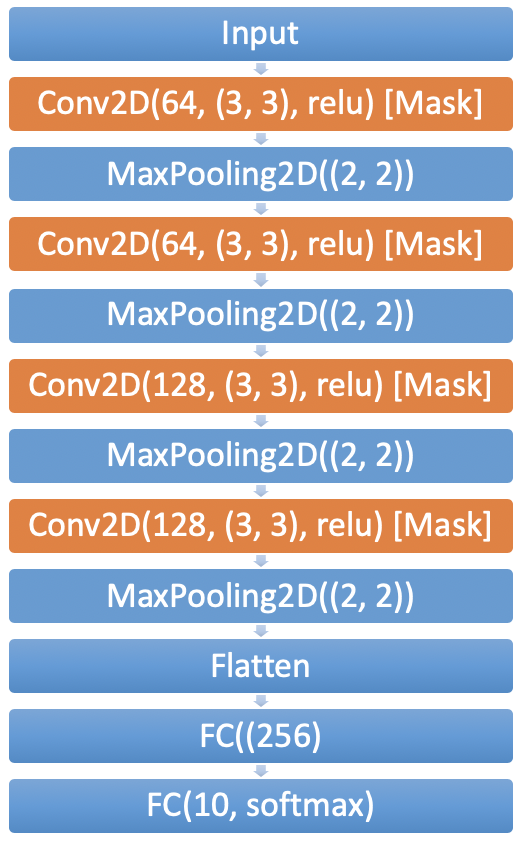}
    \centerline{Model C}
\end{minipage}
\end{center}
\captionsetup{justification=raggedright}
\vspace{0mm}
\caption{Models of the neural network architectures considered. Varying numbers of initial hidden nodes/filters are used in different experiments, but the baseline architectures remain the same. The layers where the masks are applied are written with a postfix `[Mask]', and shown in orange. The CNN models (Model B and C) are variants of the VGG architecture \cite{vgg}. 
\textit{\textbf{Model A: FC40 - FC40 (Left)}}: This is a network with two fully-connected (FC) hidden layers, each with 40 nodes. The mask is applied after each hidden layer.
\textit{\textbf{Model B: Conv64 - Conv64 (Middle)}}: This is a network with a two 2D convolutional layers each comprising of 64 filters of size 3x3 with 'same' padding. The mask is applied after the convolutional layer and before the MaxPooling2D layer.
\textit{\textbf{Model C: Conv64 - Conv64 - Conv128 - Conv128 (Right)}}: This is a network with a four 2D convolutional layers. The first two convolutional layers have 64 filters, while the next two convolutional layers have 128 filters. The filter is of size 3x3 with 'same' padding. The mask is applied after the convolutional layer and before the MaxPooling2D layer.}
\label{model_architecture}
\end{figure*}

\begin{figure*}[t]
\centering
	\begin{minipage}[t]{0.3\textwidth}
		\includegraphics[width=\textwidth]{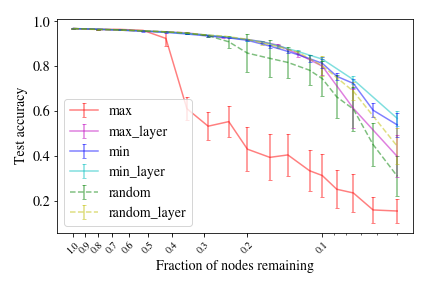}
		\caption{Test accuracy vs. fraction of nodes remaining for various metrics in Model A: FC40 - FC40 on MNIST}
		\label{fig:test_dense40_dense40}
	\end{minipage}%
	\hfill
	\begin{minipage}[t]{0.3\textwidth}
		\includegraphics[width=\textwidth]{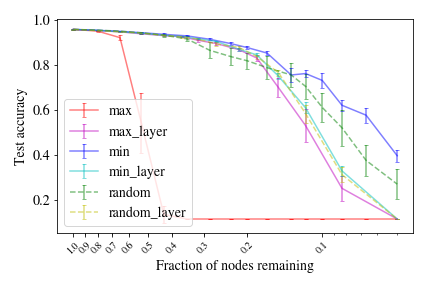}
		\caption{Test accuracy vs. fraction of nodes remaining for various metrics in Model A: FC20 - FC40 on MNIST}
		\label{fig:test_dense20_dense40}
	\end{minipage}
	\hfill
	\begin{minipage}[t]{0.3\textwidth}
		\includegraphics[width=\textwidth]{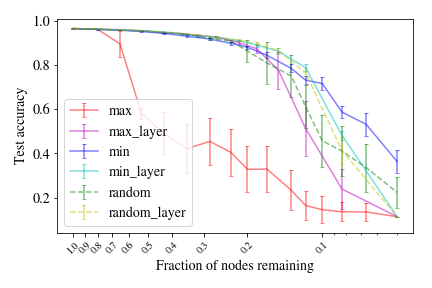}
		\caption{Test accuracy vs. fraction of nodes remaining for various metrics in Model A: FC40 - FC20 on MNIST}
		\label{fig:test_dense40_dense20}
	\end{minipage}
\end{figure*}

\begin{figure*}[t]
\centering
	\begin{minipage}[t]{0.3\textwidth}
        \includegraphics[width=\textwidth]{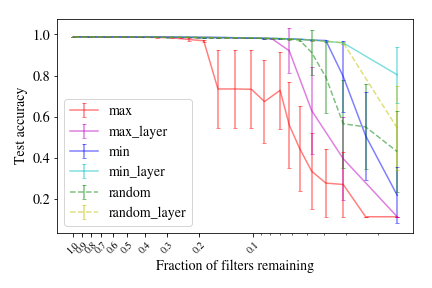}
		\caption{Test accuracy vs. fraction of filters remaining for various metrics in Model B: Conv64 - Conv64 on MNIST}
		\label{fig:test_conv64_conv64}
	\end{minipage}%
	\hfill
	\begin{minipage}[t]{0.3\textwidth}
		\includegraphics[width=\textwidth]{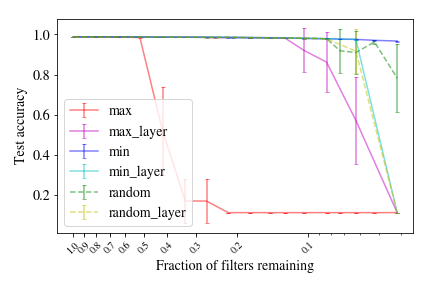}
		\caption{Test accuracy vs. fraction of filters remaining for various metrics in Model B: Conv32 - Conv64 on MNIST}
		\label{fig:test_conv32_conv64}
	\end{minipage}%
	\hfill
	\begin{minipage}[t]{0.3\textwidth}
		\includegraphics[width=\textwidth]{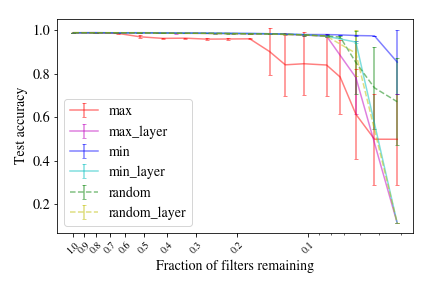}
		\caption{Test accuracy vs. fraction of filters remaining for various metrics in Model B: Conv64 - Conv32 on MNIST}
		\label{fig:test_conv64_conv32}
	\end{minipage}%
\end{figure*}

\subsection{Model Details}

The experiments are performed on a variety of network models. We use three types of feed-forward architectures: two fully-connected (FC) hidden layers of nodes (Model A), two 2D convolutional (Conv) layers (Model B) and four 2D convolutional layers (Model C). The model architectures are detailed in Fig. \ref{model_architecture}.

\section{Results}

\subsection{MLP - MNIST}
\label{MLP}

\textit{\textbf{Q1. Can DropNet perform robustly well on MLPs of various starting configurations?}}

To address this question, we conduct the experiments on different versions of Model A on MNIST, listed as follows:
\begin{enumerate}[leftmargin=*,nosep, label=1.\arabic*)]
    \item \textbf{Model A: FC40 - FC40}. The plot of test accuracy against fraction of nodes remaining for various metrics is shown in Fig. \ref{fig:test_dense40_dense40}.
    \item \textbf{Model A: FC20 - FC40}. The plot of test accuracy against fraction of nodes remaining for various metrics is shown in Fig. \ref{fig:test_dense20_dense40}.
    \item \textbf{Model A: FC40 - FC20}. The plot of test accuracy against fraction of nodes remaining for various metrics is shown in Fig. \ref{fig:test_dense40_dense20}.
\end{enumerate}

As the trends for training, validation, and test accuracies are similar, we only show the plots for test accuracy.

For 1.1), it can be seen (Fig. \ref{fig:test_dense40_dense40}) that the \texttt{minimum\_layer} performs the best, followed closely by \texttt{minimum}, then \texttt{random\_layer}, \texttt{maximum\_layer}, \texttt{random} and lastly \texttt{maximum}. The \texttt{maximum} metric performs poorly when the fraction of nodes remaining is 0.5 and below.

For 1.2), it can be seen (Fig. \ref{fig:test_dense20_dense40}) that the \texttt{minimum} metric performs the best, followed by the layer-wise and random metrics, and lastly the \texttt{maximum} metric. 

For 1.3), it can be seen (Fig. \ref{fig:test_dense40_dense20}) that the \texttt{minimum} and \texttt{minimum\_layer} are both competitive, followed by \texttt{random\_layer}, \texttt{random}, \texttt{maximum\_layer}, and lastly \texttt{maximum} metric. \texttt{minimum} performs well for all fractions of nodes remaining except between 0.1 and 0.3 where \texttt{minimum\_layer} performs slightly better. The \texttt{maximum} metric can be seen to be consistently poor when the fraction of nodes remaining is 0.6 and below.

\textbf{Evaluation:} The results indicate that, for fully connected networks, when the hidden layer sizes are equal (i.e., no bottleneck layer), the \texttt{minimum\_layer} metric is competitive. When the hidden layer sizes are unequal (i.e., there potentially exists a bottleneck layer), the \texttt{minimum} metric is competitive. This shows that \texttt{minimum} and \texttt{minimum\_layer} are competitve metrics for \textbf{Model A}. Using \textit{DropNet}, we are able to reduce the number of nodes by 60\% or more without significantly affecting model accuracy, highlighting its effectiveness in reducing network complexity. 

\subsection{CNN - MNIST}

\textit{\textbf{Q2. Can DropNet perform robustly well on CNNs of various starting configurations?}}

To address this question, we conduct experiments on different versions of Model B on MNIST, listed as follows:
\begin{enumerate}[leftmargin=*,nosep, label=2.\arabic*)]
\item \textbf{Model B: Conv64 - Conv64}. The plot of test accuracy against fraction of filters remaining for various metrics is shown in Fig. \ref{fig:test_conv64_conv64}.

\item \textbf{Model B: Conv32 - Conv64}. The plot of test accuracy against fraction of filters remaining for various metrics is shown in Fig. \ref{fig:test_conv32_conv64}.

\item \textbf{Model B: Conv64 - Conv32}. The plot of test accuracy against fraction of filters remaining for various metrics is shown in Fig. \ref{fig:test_conv64_conv32}.
\end{enumerate}

As the trends for training, validation, and test accuracies are similar, we only show the plots for test accuracy.

For 2.1), it can be seen (Fig. \ref{fig:test_conv64_conv64}) that the \texttt{minimum\_layer} metric performs the best, followed by \texttt{random\_layer}, \texttt{minimum}, \texttt{random}, \texttt{maximum\_layer}, and lastly \texttt{maximum} metric. The \texttt{maximum} metric can be seen to be consistently poor when the fraction of filters remaining is 0.3 and below.

For 2.2), it can be seen (Fig. \ref{fig:test_conv32_conv64}) that the \texttt{minimum} metric performs the best, followed by \texttt{minimum\_layer}, \texttt{random\_layer}, \texttt{random}, \texttt{maximum\_layer}, and lastly \texttt{maximum} metric. The \texttt{maximum} metric can be seen to be consistently poor when the fraction of filters remaining is 0.5 and below. The random metric is in between the performance of the \texttt{minimum} and \texttt{maximum} metrics.

For 2.3), it can be seen (Fig. \ref{fig:test_conv64_conv32}) that the \texttt{minimum} metric performs the best, followed by \texttt{minimum\_layer}, \texttt{random}, \texttt{random\_layer}, \texttt{maximum\_layer}, and lastly \texttt{maximum} metric. The \texttt{maximum} metric can be seen to be consistently poor when the fraction of filters remaining is 0.2 and below. The performance of the random metrics is in between the performance of the \texttt{minimum} and \texttt{maximum} metrics.

\textbf{Evaluation:} The findings are similar to Section \ref{MLP}. The results indicate that for convolutional layers, when the hidden layer sizes are equal (i.e., no bottleneck layer), the \texttt{minimum\_layer} metric is competitive. When the hidden layer sizes are unequal (i.e., there potentially exists a bottleneck layer), the \texttt{minimum} metric is competitive. This shows that \texttt{minimum} and \texttt{minimum\_layer} are competitive metrics for \textbf{Model B}. Using \textit{DropNet}, we are able to reduce the number of filters by 90\% or more without significantly affecting model accuracy, highlighting its effectiveness in reducing network complexity. 

\begin{figure*}[t]
\centering
	\begin{minipage}[t]{0.48\textwidth}
		\includegraphics[width=\textwidth]{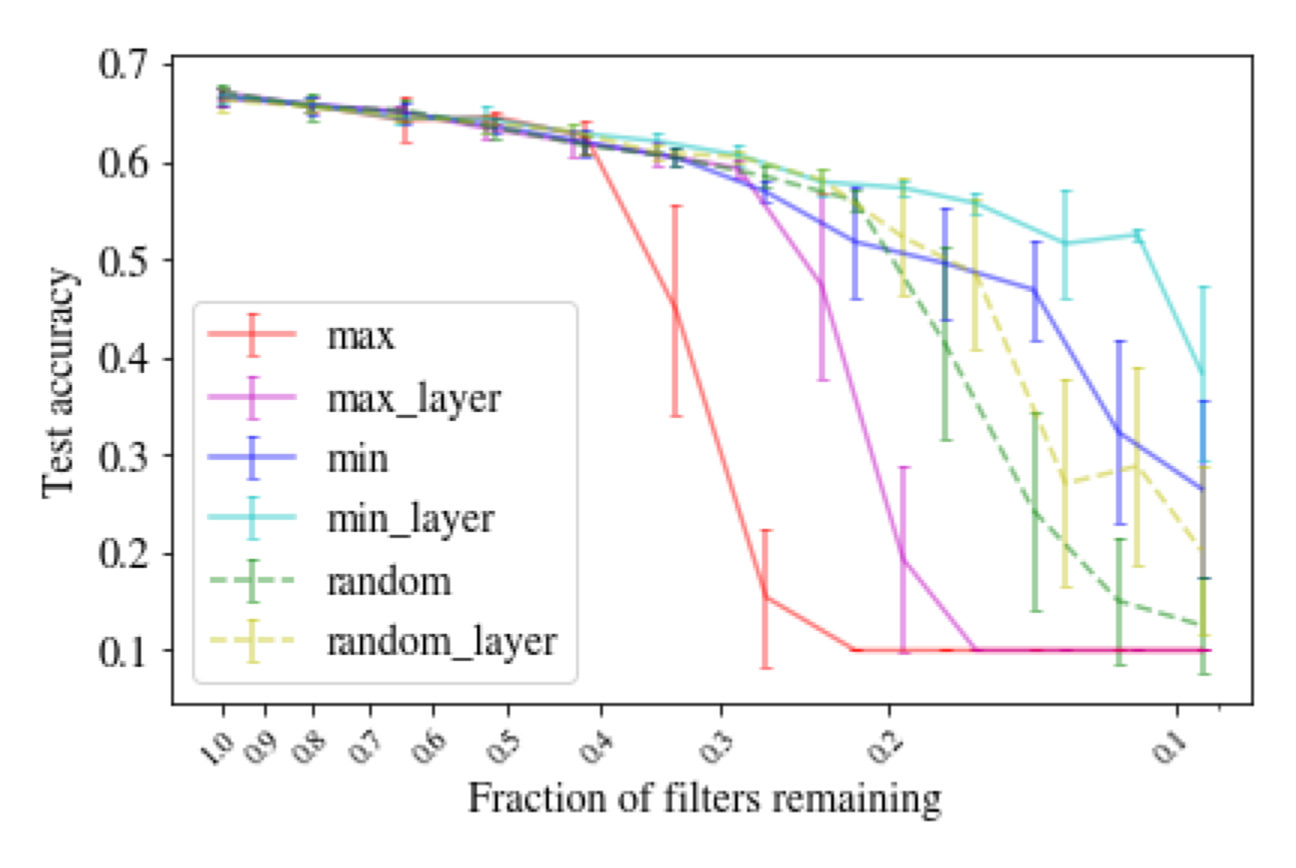}
		\caption{Test accuracy vs. fraction of filters remaining for various metrics in Model C: Conv64 - Conv64 - Conv128 - Conv128 on CIFAR-10}
		\label{fig:test_conv64x2_conv128x2}
	\end{minipage}%
	\hfill
	\begin{minipage}[t]{0.48\textwidth}
		\includegraphics[width=\textwidth]{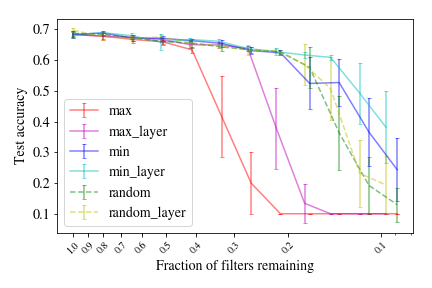}
		\caption{Test accuracy vs. fraction of filters remaining for various metrics in Model C: Conv128 - Conv128 - Conv256 - Conv256 on CIFAR-10}
		\label{fig:test_conv128x2_conv256x2}
	\end{minipage}%
\end{figure*}

\begin{figure*}[t]
\centering
	\begin{minipage}[t]{0.45\textwidth}
        \includegraphics[width=\textwidth]{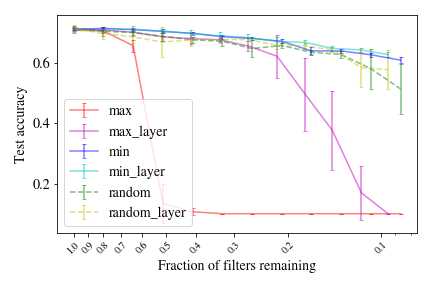}
		\caption{Test accuracy vs. fraction of filters remaining for various metrics in ResNet18 on CIFAR-10}
		\label{fig:test_resnet}
	\end{minipage}%
	\hfill
	\begin{minipage}[t]{0.45\textwidth}
		\includegraphics[width=\textwidth]{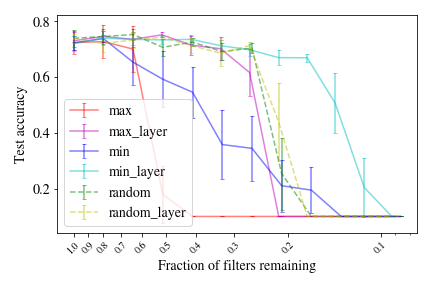}
		\caption{Test accuracy vs. fraction of filters remaining for various metrics in VGG19 on CIFAR-10}
		\label{fig:test_vgg}
	\end{minipage}%
\end{figure*}

\subsection{CNN - CIFAR-10: Model C}

\textit{\textbf{Q3. Can DropNet perform well on a larger dataset like CIFAR-10?}}

To address this question, we conduct an experiment using Model C on CIFAR-10, listed as follows:
\begin{enumerate}[leftmargin=*,nosep, label=3.\arabic*)]
\item \textbf{Model C: Conv64 - Conv64 - Conv128 - Conv128}. The plot of test accuracy against fraction of filters remaining for various metrics are shown in Fig \ref{fig:test_conv64x2_conv128x2}.
\item \textbf{Model C: Conv128 - Conv128 - Conv256 - Conv256}. The plot of training accuracy and test accuracy against fraction of filters remaining for various metrics are shown in Fig. \ref{fig:test_conv128x2_conv256x2}.
\end{enumerate}

As the trends for training, validation, and test accuracies are similar, we only show the plots for test accuracy.

It can be seen (Figs. \ref{fig:test_conv64x2_conv128x2} and  \ref{fig:test_conv128x2_conv256x2}) that the \texttt{minimum\_layer} metric performs the best, followed by \texttt{minimum}, \texttt{random\_layer}, \texttt{random}, and \texttt{maximum\_layer} and lastly \texttt{maximum} metric. The \texttt{minimum} and \texttt{minimum\_layer} perform equally well when the fraction of filters remaining is 0.5 and above. The \texttt{maximum} metric can be seen to be consistently poor when the fraction of filters remaining is 0.5 and below. The \texttt{random} metric is in between the \texttt{minimum} and \texttt{maximum} metrics.

\textbf{Evaluation:} The results show that \texttt{minimum} and \texttt{minimum\_layer} are both competitive when less than of half of the filters are dropped. Thereafter, \texttt{minimum\_layer} performs significantly better. Using \textit{DropNet}, we can reduce the number of filters by 50\% or more without significantly affecting model accuracy, highlighting its effectiveness in reducing network complexity. 

The results indicate that, for larger convolutional models like \textbf{Model C}, global pruning methods like the \texttt{minimum} metric may not be as competitive as layer-wise pruning methods like the \texttt{minimum\_layer} metric.

\subsection{CNN - CIFAR-10: ResNet18/VGG19}
\textit{\textbf{Q4. Can DropNet perform robustly well on even larger models such as ResNet18 and VGG19?}}

To address this question, we conduct an experiment using Algorithm \ref{alg:DropNet} for ResNet18 and VGG19 on CIFAR-10, both following closely to the implementation in their respective papers \cite{vgg, DBLP:journals/corr/HeZRS15}. Details of the model are in the Supplementary Material. Of note, ResNet18 is implemented without Batch Normalization, while VGG19 had a Batch Normalization before every MaxPooling2D layer. 

The plot of test accuracy against fraction of filters remaining for various metrics for ResNet18 and VGG19 is shown in Figs. \ref{fig:test_resnet} and \ref{fig:test_vgg} respectively.

For ResNet18, it can be seen (Fig. \ref{fig:test_resnet})  that the \texttt{minimum\_layer} metric performs the best, followed by \texttt{minimum}, then \texttt{random\_layer}, \texttt{random},  \texttt{maximum\_layer} and  and lastly \texttt{maximum} metric. The \texttt{minimum\_layer} and \texttt{minimum} are both competitive. 

For VGG19, it can be seen (Fig. \ref{fig:test_vgg}) that the \texttt{minimum\_layer} metric performs the best, followed by \texttt{random\_layer}, \texttt{random}, \texttt{max\_layer},  \texttt{minimum},  and  and lastly \texttt{maximum} metric. The \texttt{minimum\_layer} metric is the most competitive. 

For both ResNet18 and VGG19, the \texttt{maximum} metric can be seen to be consistently poor when the fraction of filters remaining is 0.5 and below.  The \texttt{maximum\_layer} metric is consistently poor when the fraction of filters remaining is 0.2 and below.

\textbf{Evaluation:} The results show that for larger models, \texttt{minimum\_layer} is the most competitive. It can also be seen that with the exception of \texttt{minimum} and \texttt{maximum\_layer}, the layer-wise metrics outperform the global metrics for larger models. This shows that there may be significant statistical differences between layers for larger models such that comparing magnitudes across layers may not be a good way to prune nodes/filters. That said, \texttt{minimum\_layer} can be seen to perform very well and consistently performs better than random, which shows promise that it is a good metric.

The \texttt{minimum} metric proves to be almost as competitive as \texttt{minimum\_layer} for ResNet18, but performs worse than random for VGG19. This shows that the skip connections in ResNet18 help to alleviate some of the pitfalls of global metrics. Interestingly, the \texttt{minimum} metric tends to prune out some skip connections completely, which shows that certain skip connections are unnecessary. This means that \textit{DropNet} using the \texttt{minimum} metric is able to automatically identify these redundant connections on its own.

Overall, using \textit{DropNet}, we can reduce the number of filters by 80\% or more without significantly affecting model accuracy, highlighting its effectiveness in reducing network complexity even in larger models. 

\begin{figure*}[t]
\centering
	\begin{minipage}[t]{0.48\textwidth}
		\includegraphics[width=\textwidth]{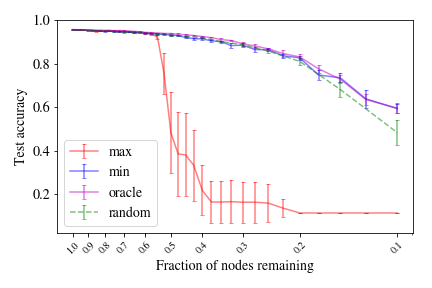}
		\caption{Test accuracy vs. fraction of nodes remaining when compared to an oracle in Model A: FC20 - FC20 on MNIST}
		\label{fig:test_oracle_dense20_dense20}
	\end{minipage}%
	\hfill
	\begin{minipage}[t]{0.48\textwidth}
		\includegraphics[width=\textwidth]{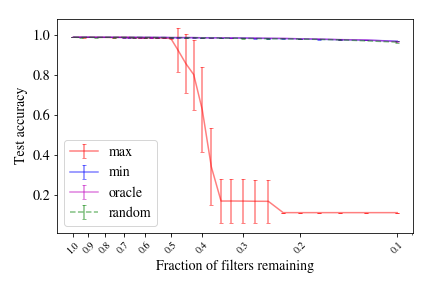}
		\caption{Test accuracy vs. fraction of filters remaining when compared to an oracle in Model B: Conv20 - Conv20 on MNIST}
		\label{fig:test_oracle_conv20_conv20}
	\end{minipage}%
\end{figure*}

\section{Empirical Analysis - Oracle Comparison}

\textit{\textbf{Q5. How competitive is the DropNet algorithm compared to an oracle?}}

There are numerous node/filter pruning methods and algorithms available, hence, rather than comparing the performance of \textit{DropNet} with these individual methods and algorithms, we utilize an \textbf{oracle} in order to establish the competitiveness of \textit{DropNet}. We define the oracle as the algorithm which \textbf{greedily} drops a node/filter out of all remaining node/filters available at every iteration of Algorithm \ref{alg:DropNet} such that the overall training loss is minimized. In order to provide a fair comparison with the oracle, nodes/filters are also pruned one at a time when using the various metrics. 

In order to perform a ``stress" test on the various metrics compared to the oracle, we analyze their performance on a smaller scale model, listed as follows:
\begin{enumerate}[leftmargin=*,nosep, label=5.\arabic*)]
\item \textbf{Model A: FC20 - FC20}. The plot of test accuracy against fraction of nodes remaining when compared against an oracle on MNIST is shown in Fig. \ref{fig:test_oracle_dense20_dense20}.

\item \textbf{Model B: Conv20 - Conv20}. The plot of test accuracy against fraction of filters remaining when compared against an oracle on MNIST is shown in Fig. \ref{fig:test_oracle_conv20_conv20}.
\end{enumerate}

For 5.1), it can be seen (Fig. \ref{fig:test_oracle_dense20_dense20}) that the oracle performs the best, followed closely by \texttt{minimum}, then \texttt{random} and lastly \texttt{maximum} metrics. The \texttt{random} metric used here is actually the \texttt{random\_layer} metric, as it provides a stronger baseline performance. Although not shown, \texttt{minimum\_layer} has similar performance to \texttt{minimum}. 

For 5.2), it can be seen (Fig. \ref{fig:test_oracle_conv20_conv20}) that the oracle, \texttt{minimum} and \texttt{random} perform equally well. The worst performing is the \texttt{maximum} metric, with poor performance with 0.4 or less fraction of filters remaining. 

\textbf{Evaluation:} The results indicate that overall, the \texttt{minimum} metric is competitive even to an oracle which minimizes training loss. This shows that the \texttt{minimum} metric is indeed a competitive criterion to drop nodes/filters.

We note that Algorithm \ref{alg:DropNet} with a pruning metric runs in linear time.
The oracle runs in polynomial time, as it has to iterate through all possible node/filter selections at the end of each iteration of Algorithm \ref{alg:DropNet}.

\begin{figure*}[t]
\centering
	\begin{minipage}[t]{0.3\textwidth}
		\includegraphics[width=\textwidth]{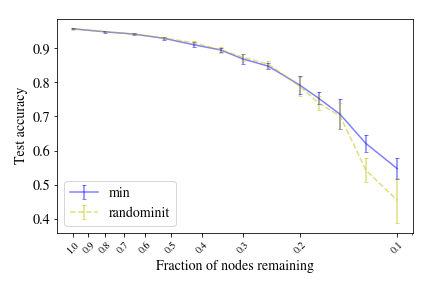}
		\caption{Test accuracy vs. fraction of nodes remaining for original initialization and random initialization in Model A: FC20 - FC20 on MNIST}
		\label{fig:test_randominit_dense20_dense20}
	\end{minipage}%
	\hfill
	\begin{minipage}[t]{0.3\textwidth}
		\includegraphics[width=\textwidth]{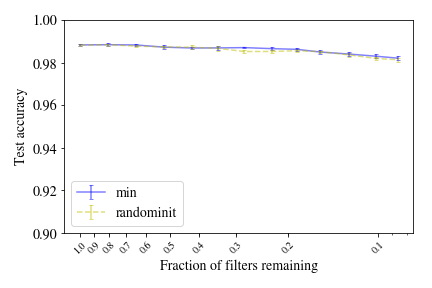}
		\caption{Test accuracy vs. fraction of filters remaining for original initialization and random initialization in Model B: Conv64 - Conv64 on MNIST}
		\label{fig:test_randominit_conv64_conv64}
	\end{minipage}%
	\hfill
	\begin{minipage}[t]{0.3\textwidth}
		\includegraphics[width=\textwidth]{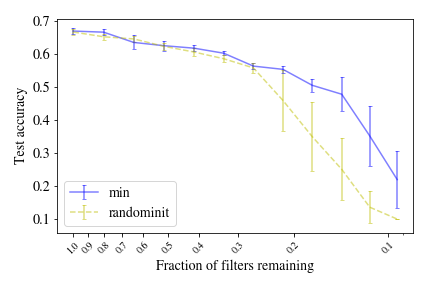}
		\caption{Test accuracy vs. fraction of nodes remaining for original init. and random init. in Model C: Conv64 - Conv64 - Conv128 - Conv128 on CIFAR-10}
		\label{fig:test_randominit_conv64x2_conv128x2}
	\end{minipage}%
\end{figure*}

\begin{figure*}[t]
\centering
	\begin{minipage}[t]{0.3\textwidth}
		\includegraphics[width=\textwidth]{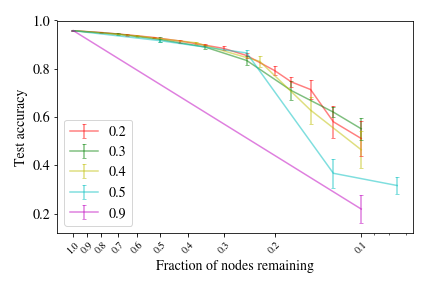}
		\caption{Test accuracy vs. fraction of filters remaining for various pruning fractions $p$ in Model A: FC20 - FC20 on MNIST}
		\label{fig:test_dense20_dense20_percent}
	\end{minipage}%
	\hfill
	\begin{minipage}[t]{0.3\textwidth}
		\includegraphics[width=\textwidth]{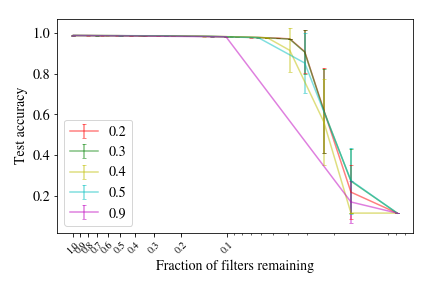}
		\caption{Test accuracy vs. fraction of filters remaining for various pruning fractions $p$ in Model B: Conv64 - Conv64 on MNIST}
		\label{fig:test_conv64x2_percent}
	\end{minipage}%
	\hfill
	\begin{minipage}[t]{0.3\textwidth}
		\includegraphics[width=\textwidth]{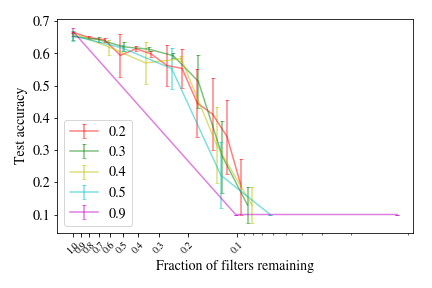}
		\caption{Test accuracy vs. fraction of filters remaining for various pruning fractions $p$ in Model C: Conv64 - Conv64 - Conv128 - Conv128 on CIFAR-10}
		\label{fig:test_conv64x2_conv128x2_percent}
	\end{minipage}%
\end{figure*}

\section{Empirical Analysis - Random Initialization}

\textit{\textbf{Q6. Is the starting initialization of weights and biases important?}}

We compare the performance of a network retaining its initial weights and biases $\theta_0$ when performing iterative node/filter pruning, as compared to a network with the pruned architecture but with a random initialization (\texttt{randominit}). We conduct the following experiments:

\begin{enumerate}[leftmargin=*,nosep, label=6.\arabic*)]
\item \textbf{Model A: FC20 - FC20}. The plot of test accuracy against fraction of nodes remaining for original and random initialization is shown in Fig. \ref{fig:test_randominit_dense20_dense20}. 
\item \textbf{Model B: Conv64 - Conv64}. The plot of test accuracy against fraction of filters remaining for original and random initialization is shown in Fig. \ref{fig:test_randominit_conv64_conv64}.
\item \textbf{Model C: Conv64 - Conv64 - Conv128 - Conv128}. The plot of test accuracy against fraction of filters remaining for original and random initialization is shown in Fig. \ref{fig:test_randominit_conv64x2_conv128x2}.
\end{enumerate}

It can be seen (Figs. \ref{fig:test_randominit_dense20_dense20}, \ref{fig:test_randominit_conv64_conv64}, and  \ref{fig:test_randominit_conv64x2_conv128x2}) that unlike the Lottery Ticket Hypothesis (see Figure 4 in \cite{Frankle2018}), \textit{DropNet} does not suffer from loss of performance when randomly initialized for up to 70\% to 80\% of the nodes/filters being dropped.

\textbf{Evaluation:} This means that for \textit{DropNet}, only the final pruned network architecture is important, and not the initial weights and biases of the network. One reason that the original initialization is not important may be because the learning rate is high enough (0.1) for the network to retrain sufficiently given just the model architecture. This concurs with the finding that the 'winning ticket' in the Lottery Ticket Hypothesis does not confer significant advantages over random reinitialization if a larger learning rate of 0.1 is used instead of 0.01 \cite{liu2018rethinking}.

Similar results are also obtained for larger models such as ResNet18 and VGG19 (details in Supplementary Material), which shows that this finding is a general one.

\section{Empirical Analysis - Percentage of nodes/filters to Drop}

\textit{\textbf{Q7. Can we drop more nodes/filters at a time to reduce number of training cycles and prune the model faster without affecting accuracy?}}

We explore this question by analyzing the performance of the \texttt{minimum} metric. We compare the performance of the model using different values of the pruning fraction $p = 0.2, 0.3, 0.4, 0.5$ and $0.9$. The experiments performed are as follows:

\begin{enumerate}[leftmargin=*,nosep, label=7.\arabic*)]
\item \textbf{Model A: FC20 - FC20}. The plot of test accuracy against fraction of nodes remaining for various pruning fractions $p$ on MNIST is shown in Fig. \ref{fig:test_dense20_dense20_percent}.
\item \textbf{Model B: Conv64 - Conv64}. The plot of test accuracy against fraction of filters remaining for various pruning fractions $p$ on MNIST is shown in Fig. \ref{fig:test_conv64x2_percent}.
\item \textbf{Model C: Conv64 - Conv64 - Conv128 - Conv128}. The plot of test accuracy against fraction of nodes remaining for various pruning fractions $p$ on CIFAR-10 is shown in Fig. \ref{fig:test_conv64x2_conv128x2_percent}.
\end{enumerate}

\textbf{Evaluation:} Overall, it can be seen (Figs. \ref{fig:test_dense20_dense20_percent}, \ref{fig:test_conv64x2_percent}, and  \ref{fig:test_conv64x2_conv128x2_percent}) that dropping a greater fraction $p$ of nodes/filters per training cycle leads to poorer performance. For one-shot pruning methods which attempt to remove 90\% of nodes/filters, it is not ideal as it generally leads to worse performance. This further reinforces the competitiveness of iterative pruning as compared to one-shot pruning. 

The results also show that larger $p$ has similar performance when dropping up to 70\% of the nodes/filters. Hence, it may be possible to iterate faster through Algorithm \ref{alg:DropNet} by adopting a larger $p$ under certain conditions. Further experiments need to be done to determine the optimal pruning fraction $p$, but $p = 0.2$ seems to be competitive.

\section{Concluding Remarks}

{\bf Reflections:}
In this paper, we propose \textit{DropNet}, which iteratively drops nodes/filters and, hence, reduces network complexity. We illustrate how \textit{DropNet} can potentially reduce network size by up to 90\% without any significant loss of accuracy. Also, there does not need to be a particular initialization required when dropping up to 70\% of the nodes/filters. \textit{DropNet} also has similar performance to an oracle which greedily removes nodes/filters one at a time to minimise training loss, which shows its competitiveness.

\textit{DropNet}, utilizing either the \texttt{minimum} or \texttt{minimum\_layer} metric, proves to be competitive over a variety of model architectures. The \texttt{minimum} metric seems to work robustly well for smaller models such as Models A and B, while the \texttt{minimum\_layer} metric appears to be better for larger models such as Model C, ResNet18 and VGG19, in the configurations we considered. 

We conjecture that one reason for the better performance of the \texttt{minimum\_layer} metric in larger models is that as the number of layers increases, the statistical properties of the post-activation values of the nodes/filters may be significantly different in each layer. Hence, using a single metric to prune globally may not be as good as pruning layer-wise. The exception is when using skip connections, as empirical results suggest that the \texttt{minimum} metric is also competitive for larger models for ResNet18. This seems to suggest that skip connections work well with \textit{DropNet}.

{\bf Additional Experiments:} In our Supplementary Material, we show that DropNet is scalable and achieves similar results on Tiny ImageNet . Furthermore, we also compare \textit{DropNet} with another data-driven approach, APoZ, and show that \textit{DropNet} outperforms APoZ and can achieve better test accuracy for the same amount of pruning.

{\bf Future Work:} To further show \textit{DropNet}'s generalizability, more experiments can be done on i) alternative neural network architectures such as RNNs, as well as ii) other domains such as NLP and reinforcement learning. \textit{DropNet} has been shown to work well empirically with $ReLU$ activation functions, and it remains to be seen whether other metrics may be required for other activation functions such as $sigmoid$, $tanh$ and $ReLU$ variants.

{\bf Source Code:} To encourage further research on iterative pruning techniques, the source code used for our experiments is publicly available at \href{https://github.com/tanchongmin/DropNet}{https://github.com/tanchongmin/DropNet}.

\newpage
{\bf \large Acknowledgements}

\textcolor{black}{
This research is supported by the National Research Foundation, Singapore under its AI Singapore Programme (AISG Award No: AISG-GC-2019-002). Any opinions, findings and conclusions or recommendations expressed in this material are those of the author(s) and do not reflect the views of National Research Foundation, Singapore.}

%
%

\bibliography{DropNet.bib}
\bibliographystyle{icml2020}

\newpage
\setcounter{table}{0}
\renewcommand{\thetable}{\Alph{section}\arabic{table}}
\setcounter{figure}{0}
\renewcommand{\thefigure}{\Alph{section}\arabic{figure}}
\appendix

\section{Summary (Supplementary Materials)}
\begin{figure*}[!t]
\begin{center}
\hfill
\begin{minipage}[t]{0.45\textwidth}
\centering
    \includegraphics[height=3in]{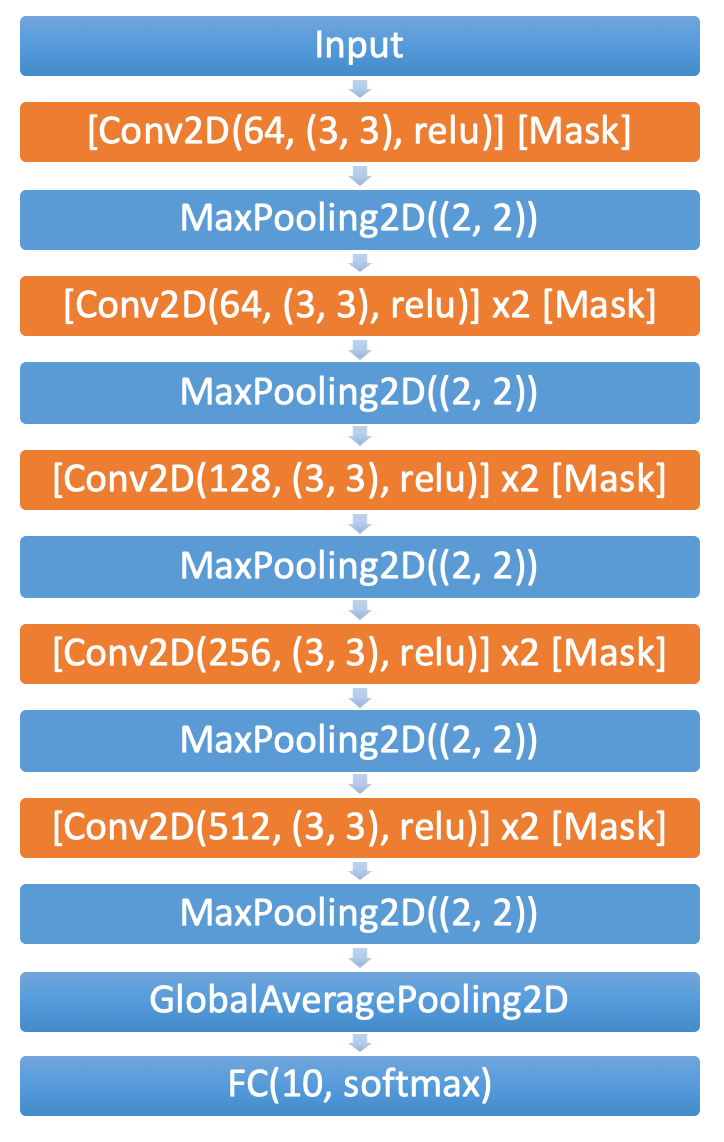}
    \centerline{ResNet18}
\end{minipage}
\hfill
\begin{minipage}[t]{0.45\textwidth}
\centering
    \includegraphics[height=3in]{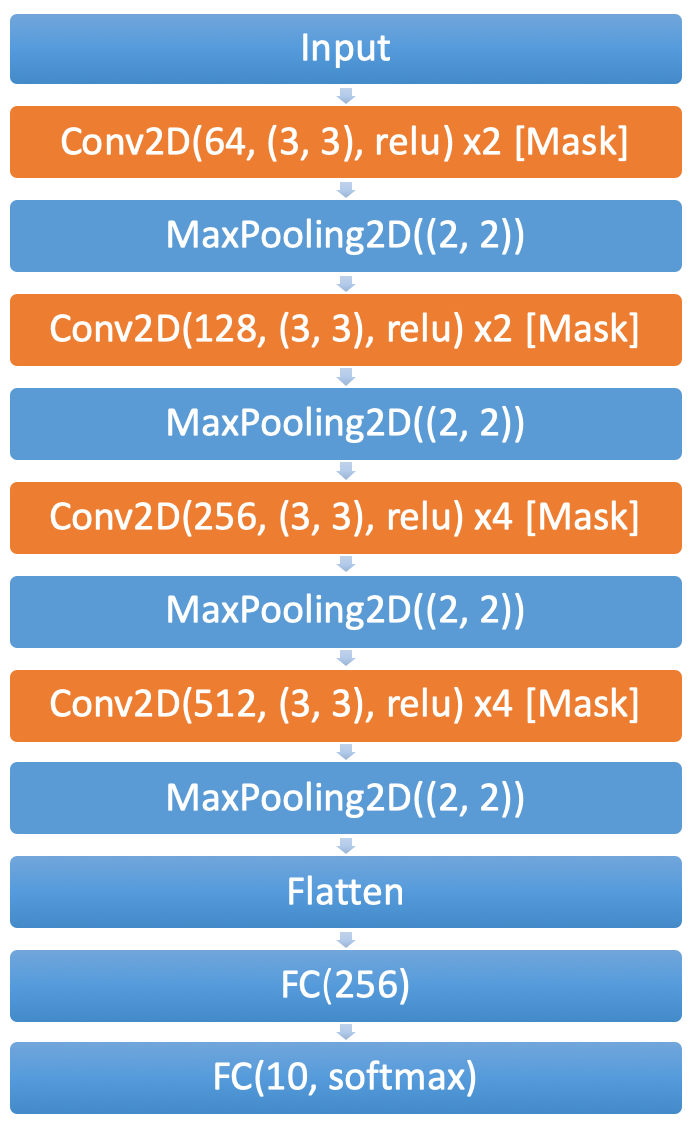}
    \centerline{VGG19}
\end{minipage}
\end{center}
\captionsetup{justification=raggedright}
\vspace{0mm}
\caption{Architecture of ResNet18 \cite{DBLP:journals/corr/HeZRS15} and VGG19 \cite{vgg} used in the experiments. The layers where the masks are applied are written with a postfix `[Mask]', and shown in orange.
\textit{\textbf{ResNet18 (Left)}}: The network architecture closely follows that of ResNet18. It consists of several skip-connection blocks which are shown in square brackets. The model consists of repeated residual blocks comprising two 2D convolutional layers followed by a MaxPooling2D layer of stride 2. Each 2D convolutional layer comprises either 64, 128, 256 or 512 filters, of size 3x3 with 'same' padding. After the multiple residual blocks, the filters are averaged using GlobalAveragePooling2D before passing into the final fully connected layer with 10 nodes. The mask is applied after the convolutional layer and before the MaxPooling2D layer. Batch Normalization is not applied between layers as the model is found to work well even without it.
\textit{\textbf{VGG19 (Right)}}: This is a network with repeated blocks of 2/4 2D convolutional layers followed by a MaxPooling2D layer. The 2D convolutional layer comprises either 64, 128, 256 or 512 filters, of size 3x3 with 'same' padding. The mask is applied after the convolutional layer and before the MaxPooling2D layer. Batch normalization is applied right before every MaxPooling2D layer.}
\label{large_model_architecture}
\end{figure*}

\begin{figure*}[t]
\centering
	\begin{minipage}[t]{0.3\textwidth}
        \includegraphics[width=\textwidth]{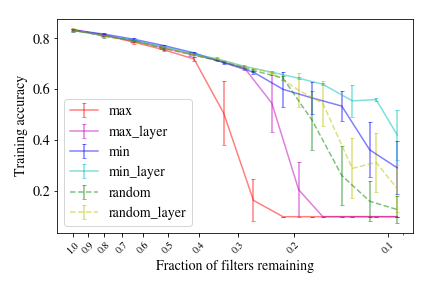}
		\caption{Plot of training accuracy against fraction of filters remaining for various metrics in Model C: Conv64 - Conv64 - Conv128 - Conv128 on CIFAR-10}
		\label{fig:train_conv64x2_conv128x2}
	\end{minipage}%
	\hfill
	\begin{minipage}[t]{0.3\textwidth}
		\includegraphics[width=\textwidth]{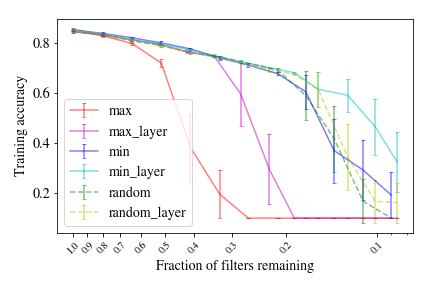}
		\caption{Plot of training accuracy against fraction of filters remaining for various metrics in Model C: Conv128 - Conv128 - Conv128 - Conv128 on CIFAR-10}
		\label{fig:train_conv128x2_conv128x2}
	\end{minipage}%
	\hfill
	\begin{minipage}[t]{0.3\textwidth}
		\includegraphics[width=\textwidth]{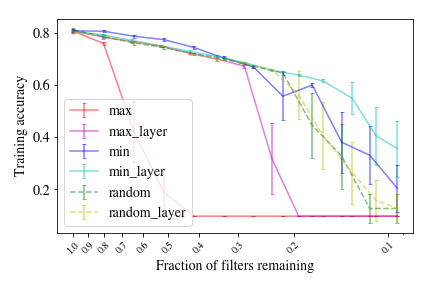}
		\caption{Plot of training accuracy against fraction of filters remaining for various metrics in Model C: Conv128 - Conv128 - Conv64 - Conv64 on CIFAR-10}
		\label{fig:train_conv128x2_conv64x2}
	\end{minipage}%
\end{figure*}

\begin{figure*}[t]
\centering
	\begin{minipage}[t]{0.3\textwidth}
        \includegraphics[width=\textwidth]{Supplementary 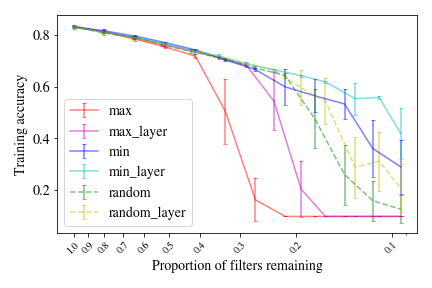}
		\caption{Plot of test accuracy against fraction of filters remaining for various metrics in Model C: Conv64 - Conv64 - Conv128 - Conv128 on CIFAR-10}
		\label{fig:test_conv64x2_conv128x2}
	\end{minipage}%
	\hfill
	\begin{minipage}[t]{0.3\textwidth}
		\includegraphics[width=\textwidth]{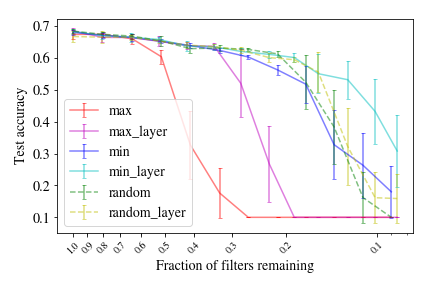}
		\caption{Plot of test accuracy against fraction of filters remaining for various metrics in Model C: Conv128 - Conv128 - Conv128 - Conv128 on CIFAR-10}
		\label{fig:test_conv128x2_conv128x2}
	\end{minipage}%
	\hfill
	\begin{minipage}[t]{0.3\textwidth}
		\includegraphics[width=\textwidth]{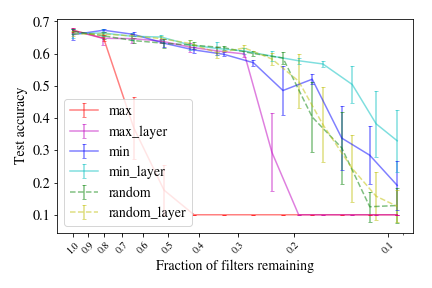}
		\caption{Plot of test accuracy against fraction of filters remaining for various metrics in Model C: Conv128 - Conv128 - Conv64 - Conv64 on CIFAR-10}
		\label{fig:test_conv128x2_conv64x2}
	\end{minipage}%
\end{figure*}

In this supplementary material, we present the following:

\begin{enumerate}
\item Results using more variants of Model C (Figs. \ref{fig:train_conv64x2_conv128x2},  \ref{fig:train_conv128x2_conv128x2}, \ref{fig:train_conv128x2_conv64x2}, \ref{fig:test_conv64x2_conv128x2}, \ref{fig:test_conv128x2_conv128x2},  \ref{fig:test_conv128x2_conv64x2}) on the CIFAR-10 dataset 

\item Results using ResNet18 (Figs. \ref{fig:train_resnet}, \ref{fig:test_resnet}) and VGG19 (Figs. \ref{fig:train_vgg}, \ref{fig:test_vgg}) on the CIFAR-10 dataset

\item Results of random initialization of ResNet18 (Figs. \ref{fig:test_resnet_randominit}, \ref{fig:test_resnet_randominit_layer}) and VGG19 (Figs. \ref{fig:test_vgg_randominit}, \ref{fig:test_vgg_randominit_layer}) on the CIFAR-10 dataset

\item Results using ResNet18 (Figs. \ref{fig:train_resnet_tinyimagenet}, \ref{fig:test_resnet_tinyimagenet}) and VGG19 (Figs. \ref{fig:train_vgg_tinyimagenet}, \ref{fig:test_vgg_tinyimagenet}) on the Tiny ImageNet dataset 

\item Performance comparison to Average Percentage of Zeros (APoZ) \cite{hu2016data} for ResNet18 (Figs. \ref{fig:train_resnet_apoz}, \ref{fig:test_resnet_apoz}) and VGG19 (Figs. \ref{fig:train_vgg_apoz}, \ref{fig:test_vgg_apoz}) on the CIFAR-10 dataset

\end{enumerate}

The results show that \textit{DropNet} is robust for larger models, and the final pruned model is able to achieve a similar performance even after reinitialization. DropNet also has better empirical performance than prior data-driven approach APoZ and is able to achieve better test accuracy for the same amount of pruning.

\section{Methodology}

The supplementary experiments performed use the same methodology as the main paper. In addition, to show \textit{DropNet}'s scalability, we also perform experiments on the Tiny ImageNet dataset. We demonstrate how effective pruning using \textit{DropNet} can be done on larger models like Model C (Conv4), ResNet18 and VGG19. For ResNet18 and VGG19, the model architecture follows closely from the original papers \cite{vgg, DBLP:journals/corr/HeZRS15} and are detailed in Fig. \ref{large_model_architecture}.

Algorithm 1, which is used throughout the supplementary material, is detailed in the main paper.

\section {Experiments}
\subsection{CNN - CIFAR-10: Model C (Conv4)}
\label{section:CIFARbase}

\textit{\textbf{Q1. Can DropNet perform robustly well on larger CNNs of various starting configurations?}}

To address this question, we conduct an experiment using Algorithm 1 for various configurations of Model C on CIFAR-10, listed as follows:
\begin{enumerate}[leftmargin=*,nosep, label=1.\arabic*)]
\item \textbf{Model C: Conv64 - Conv64 - Conv128 - Conv128}. The plot of training and test accuracy against fraction of filters remaining for various metrics are shown in Figs. \ref{fig:train_conv64x2_conv128x2} and \ref{fig:test_conv64x2_conv128x2} respectively.
\item \textbf{Model C: Conv128 - Conv128 - Conv128 - Conv128}. The plot of training accuracy and test accuracy against fraction of filters remaining for various metrics are shown in Figs. \ref{fig:train_conv128x2_conv128x2} and \ref{fig:test_conv128x2_conv128x2} respectively.
\item \textbf{Model C: Conv128 - Conv128 - Conv64 - Conv64}. The plot of training accuracy and test accuracy against fraction of filters remaining for various metrics are shown in Figs. \ref{fig:train_conv128x2_conv64x2} and \ref{fig:test_conv128x2_conv64x2} respectively.
\end{enumerate}

For 1.1), it can be seen (Figs. \ref{fig:train_conv64x2_conv128x2} and  \ref{fig:test_conv64x2_conv128x2}) that the \texttt{minimum\_layer} metric performs the best, followed by \texttt{minimum}, \texttt{random\_layer}, \texttt{random}, and \texttt{maximum\_layer} and lastly \texttt{maximum} metric. The \texttt{minimum} and \texttt{minimum\_layer} perform equally well when the fraction of filters remaining is 0.3 and above. The \texttt{maximum} metric can be seen to be consistently poor when the fraction of filters remaining is 0.5 and below. The \texttt{random} metric is in between the performance of the \texttt{minimum} and \texttt{maximum} metrics.

For 1.2), it can be seen (Figs. \ref{fig:train_conv128x2_conv128x2} and  \ref{fig:test_conv128x2_conv128x2}) that the \texttt{minimum\_layer} metric performs the best, followed by  \texttt{random\_layer},  \texttt{random}, \texttt{minimum},  \texttt{maximum\_layer} and lastly \texttt{maximum} metric. The \texttt{minimum} and \texttt{minimum\_layer} perform equally well when the fraction of filters remaining is 0.3 and above. The \texttt{maximum} metric can be seen to be consistently poor when the fraction of filters remaining is 0.6 and below.

For 1.3), it can be seen (Figs. \ref{fig:train_conv128x2_conv64x2} and  \ref{fig:test_conv128x2_conv64x2}) that the \texttt{minimum\_layer} metric performs the best, followed by \texttt{minimum}, \texttt{random\_layer}, \texttt{random}, and \texttt{maximum\_layer} and lastly \texttt{maximum} metric. The \texttt{minimum} and \texttt{minimum\_layer} perform equally well when the fraction of filters remaining is 0.4 and above, with the \texttt{minimum} metric performing significantly better when the fraction of filters remaining is 0.6 and above, even outperforming the original model accuracy at some instances. The \texttt{maximum} metric can be seen to be consistently poor when the fraction of filters remaining is 0.8 and below. 

\textbf{Evaluation:} The results show that \texttt{minimum} and \texttt{minimum\_layer} are both competitive when less than of half of the filters are dropped. Thereafter, \texttt{minimum\_layer} performs significantly better. Using \textit{DropNet}, we can reduce the number of filters by 50\% or more without significantly affecting model accuracy, highlighting its effectiveness in reducing network complexity. 

The results indicate that, for larger convolutional models like Model C, global pruning methods like the \texttt{minimum} metric are only good at the early stages of pruning. In fact, for models with non-symmetric layers (see Figs. \ref{fig:train_conv128x2_conv64x2} and \ref{fig:test_conv128x2_conv64x2}), \texttt{minimum} works the best when the fraction of filters remaining is 0.5 and above, and may even outperform the original model's accuracy. We posit that this is due to the flexibility of global pruning methods to avoid pruning small layers which pose a bottleneck as compared to layer-wise pruning methods. That said, not all global pruning methods can do that - \texttt{maximum} and \texttt{random} do not display such a trend of avoiding bottlenecks.

One further observation is that the train and test data show similar accuracy trends (the same applies for validation accuracy, although not shown here). This shows that the train-test-validation split is done well and the general distribution of the train dataset is similar to that of the test dataset. Hence, a metric to prune based on the node's post-activation value such as \textit{DropNet} works well using just the post-activation values from the training data only.

\begin{figure*}[t]
\centering
	\begin{minipage}[t]{0.45\textwidth}
        \includegraphics[width=\textwidth]{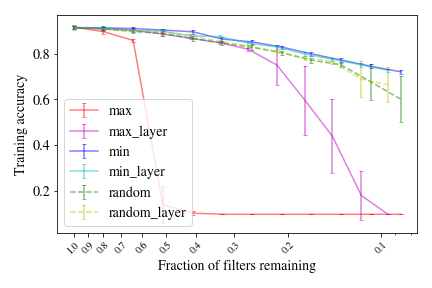}
		\caption{Plot of training accuracy against fraction of filters remaining for various metrics in ResNet18 on CIFAR-10}
		\label{fig:train_resnet}
	\end{minipage}%
	\hfill
	\begin{minipage}[t]{0.45\textwidth}
		\includegraphics[width=\textwidth]{Supplementary Images/test_accuracy_evaluate_resnet18.png}
		\caption{Plot of test accuracy against fraction of filters remaining for various metrics in ResNet18 on CIFAR-10}
		\label{fig:test_resnet}
	\end{minipage}%
\end{figure*}

\begin{figure*}[t]
\centering
	\begin{minipage}[t]{0.45\textwidth}
        \includegraphics[width=\textwidth]{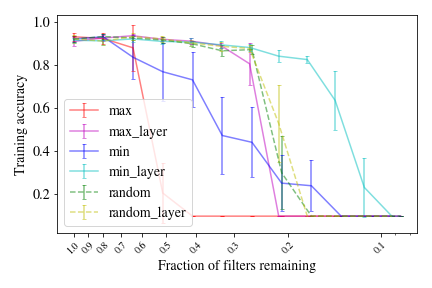}
		\caption{Plot of training accuracy against fraction of filters remaining for various metrics in VGG19 on CIFAR-10}
		\label{fig:train_vgg}
	\end{minipage}%
	\hfill
	\begin{minipage}[t]{0.45\textwidth}
		\includegraphics[width=\textwidth]{Supplementary Images/test_accuracy_evaluate_vgg.png}
		\caption{Plot of test accuracy against fraction of filters remaining for various metrics in VGG19 on CIFAR-10}
		\label{fig:test_vgg}
	\end{minipage}%
\end{figure*}

\begin{figure*}[t]
\centering
	\begin{minipage}[t]{0.45\textwidth}
        \includegraphics[width=\textwidth]{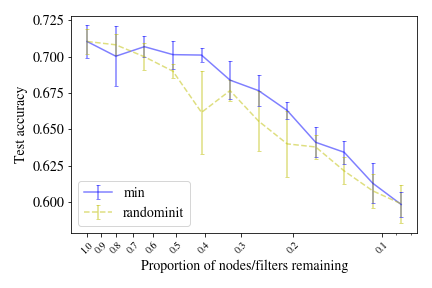}
		\caption{Plot of test accuracy against fraction of filters remaining for original initialization and random initialization in ResNet18 using pruned model from \texttt{minimum} metric on CIFAR-10}
		\label{fig:test_resnet_randominit}
	\end{minipage}%
	\hfill
	\begin{minipage}[t]{0.45\textwidth}
		\includegraphics[width=\textwidth]{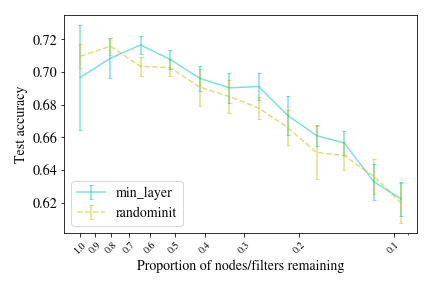}
		\caption{Plot of test accuracy against fraction of filters remaining for original initialization and random initialization in ResNet18 using pruned model from \texttt{mininum\_layer} metric on CIFAR-10}
		\label{fig:test_resnet_randominit_layer}
	\end{minipage}%
\end{figure*}

\begin{figure*}[t]
\centering
    \begin{minipage}[t]{0.45\textwidth}
		\includegraphics[width=\textwidth]{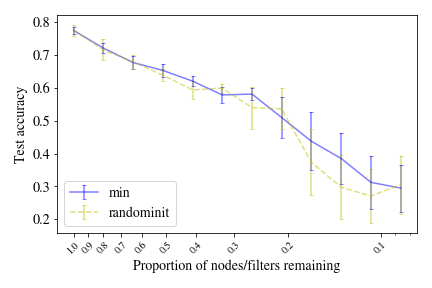}
		\caption{Plot of test accuracy against fraction of filters remaining for original initialization and random initialization in VGG19 using pruned model from \texttt{minimum} metric on CIFAR-10}
		\label{fig:test_vgg_randominit}
	\end{minipage}%
	\hfill
	\begin{minipage}[t]{0.45\textwidth}
        \includegraphics[width=\textwidth]{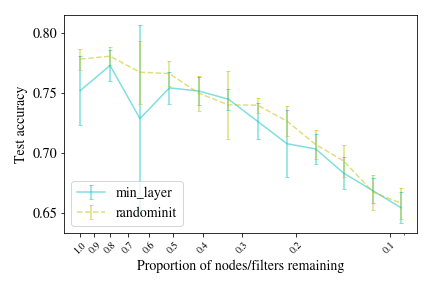}
		\caption{Plot of test accuracy against fraction of filters remaining for original initialization and random initialization in VGG19 using pruned model from \texttt{minimum\_layer} metric on CIFAR-10}
		\label{fig:test_vgg_randominit_layer}
	\end{minipage}%
    
\end{figure*}

\subsection{CNN - CIFAR-10: ResNet18/VGG19}
\textit{\textbf{Q2. Can DropNet perform robustly well on even larger models such as ResNet18 and VGG19?}}

To address this question, we conduct an experiment using Algorithm 1 for ResNet18 and VGG19 on CIFAR-10:

\begin{enumerate}[leftmargin=*,nosep, label=2.\arabic*)]
\item \textbf{ResNet18}. The plot of training and test accuracy against fraction of filters remaining for various metrics is shown in Figs. \ref{fig:train_resnet} and \ref{fig:test_resnet} respectively.
\item \textbf{VGG19}. The plot of training and test accuracy against fraction of filters remaining for various metrics is shown in Figs. \ref{fig:train_vgg} and \ref{fig:test_vgg} respectively.
\end{enumerate}

For ResNet18, it can be seen (Figs. \ref{fig:train_resnet} and \ref{fig:test_resnet})  that the \texttt{minimum\_layer} metric performs the best, followed by \texttt{minimum}, then \texttt{random\_layer}, \texttt{random},  \texttt{maximum\_layer} and  and lastly \texttt{maximum} metric. The \texttt{minimum\_layer} and \texttt{minimum} are both competitive. 

For VGG19, it can be seen (Figs. \ref{fig:train_vgg} and \ref{fig:test_vgg}) that the \texttt{minimum\_layer} metric performs the best, followed by \texttt{random\_layer}, \texttt{random}, \texttt{max\_layer},  \texttt{minimum},  and  and lastly \texttt{maximum} metric. The \texttt{minimum\_layer} metric is the most competitive. 

For both ResNet18 and VGG19,  \texttt{maximum} can be seen to be consistently poor when the fraction of filters remaining is 0.5 and below, while \texttt{maximum\_layer} is consistently poor when the fraction of filters remaining is 0.2 and below.

\textbf{Evaluation:} The results show that for larger models, \texttt{minimum\_layer} is the most competitive. It can also be seen that with the exception of \texttt{minimum} and \texttt{maximum\_layer}, the layer-wise metrics outperform the global metrics for larger models. This shows that there may be significant statistical differences between layers for larger models such that comparing magnitudes across layers may not be a good way to prune nodes/filters. That said, the \texttt{minimum\_layer} can be seen to perform very well and consistently performs better than random, which shows promise that it is a good metric.

The \texttt{minimum} metric proves to be almost as competitive as \texttt{minimum\_layer} for ResNet18, but performs worse than random for VGG19. This shows that the skip connections in ResNet18 does help to alleviate some of the pitfalls of global metrics. Interestingly, the \texttt{minimum} metric tends to prune out some skip connections completely, which shows that certain skip connections are unnecessary. This means that \textit{DropNet} using the \texttt{minimum} metric is able to automatically identify these redundant connections on its own.

In comparison, it can be seen that the \texttt{maximum} metric performs the worse in all cases, and shows that filters with high expected absolute post-activate values are generally important in classification and should not be removed. 

The \texttt{maximum\_layer} metric on the other hand, performs poorly in ResNet18, but has comparable performance to the layer-wise metrics in VGG19. This may be due to the fact that VGG19 in the experiments use a Batch Normalization after every change of Conv2D filter size, which helps to normalize the post-activation values and hence, the layer-wise pruning metrics do not differ much in performance.

Using \textit{DropNet}, we can reduce the number of filters by 80\% or more without significantly affecting model accuracy, highlighting its effectiveness in reducing network complexity. 

\subsection{CNN - CIFAR-10: ResNet18/VGG19 (Random Initialization)}

\textit{\textbf{Q3. Is the starting initialization of weights and biases important for larger models such as ResNet18 and VGG19?}}

We compare the performance of a network retaining its initial weights and biases $\theta_0$ when performing iterative node/filter pruning, as compared to a network with the pruned architecture but with a random initialization (\texttt{randominit}). In our experiments, we focus on the pruned architecture produced by \textit{DropNet} metrics, namely \texttt{minimum} and \texttt{minimum\_layer}. The experiments are conducted on CIFAR-10, and are detailed as follows:

\begin{enumerate}[leftmargin=*,nosep, label=3.\arabic*)]
\item \textbf{ResNet18}. The plot of test accuracy against fraction of nodes remaining for original and random initialization using pruned model from \texttt{minimum} metric and \texttt{minimum\_layer} metric respectively are shown in Figs. \ref{fig:test_resnet_randominit} and \ref{fig:test_resnet_randominit_layer}. 
\item \textbf{VGG19}. The plot of test accuracy against fraction of nodes remaining for original and random initialization using pruned model from \texttt{minimum} metric and \texttt{minimum\_layer} metric respectively are shown in Figs. \ref{fig:test_vgg_randominit} and \ref{fig:test_vgg_randominit_layer}. 
\end{enumerate}

It can be seen (Figs. \ref{fig:test_resnet_randominit}, \ref{fig:test_resnet_randominit_layer},  \ref{fig:test_vgg_randominit}, \ref{fig:test_vgg_randominit_layer}) that unlike the Lottery Ticket Hypothesis (see Figure 4 in \cite{Frankle2018}), \textit{DropNet} does not suffer from loss of performance when randomly initialized.

\textbf{Evaluation:} This means that for \textit{DropNet}, only the final pruned network architecture is important, and not the initial weights and biases of the network. This is a pleasant finding as it shows that \textit{DropNet} can prune a model down to an ideal structure, from which it can be readily deployed on modern machine learning libraries.

\begin{figure*}[t]
\centering
	\begin{minipage}[t]{0.45\textwidth}
        \includegraphics[width=\textwidth]{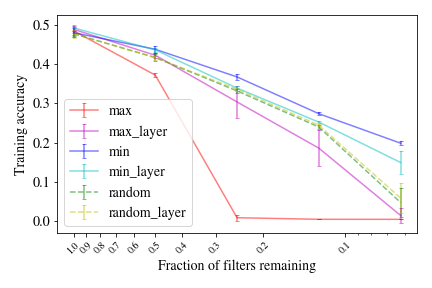}
		\caption{Plot of training accuracy against fraction of filters remaining for various metrics in ResNet18 on TinyImageNet}
		\label{fig:train_resnet_tinyimagenet}
	\end{minipage}%
	\hfill
	\begin{minipage}[t]{0.45\textwidth}
		\includegraphics[width=\textwidth]{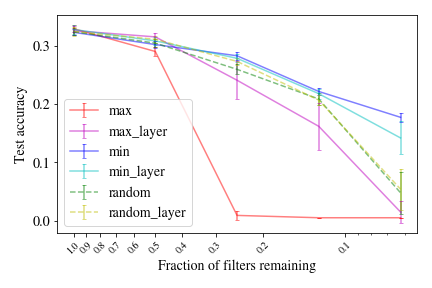}
		\caption{Plot of test accuracy against fraction of filters remaining for various metrics in ResNet18 on TinyImageNet}
		\label{fig:test_resnet_tinyimagenet}
	\end{minipage}%
\end{figure*}

\begin{figure*}[t]
\centering
	\begin{minipage}[t]{0.45\textwidth}
        \includegraphics[width=\textwidth]{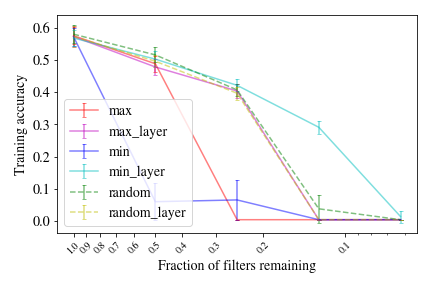}
		\caption{Plot of training accuracy against fraction of filters remaining for various metrics in VGG19 on TinyImageNet}
		\label{fig:train_vgg_tinyimagenet}
	\end{minipage}%
	\hfill
	\begin{minipage}[t]{0.45\textwidth}
		\includegraphics[width=\textwidth]{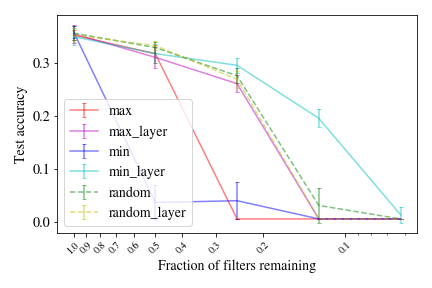}
		\caption{Plot of test accuracy against fraction of filters remaining for various metrics in VGG19 on TinyImageNet}
		\label{fig:test_vgg_tinyimagenet}
	\end{minipage}%
\end{figure*}

\subsection{CNN - Tiny ImageNet: ResNet18/VGG19}

\textit{\textbf{Q4. Can DropNet perform well for even larger datasets such as Tiny ImageNet?}}

In order to show the generalizability of the \textit{DropNet} algorithm on larger datasets, we utilize the Tiny ImageNet dataset, which is a smaller-resolution parallel of the larger ImageNet dataset. This dataset was taken from \url{https://tiny-imagenet.herokuapp.com/}.

\textbf{Dataset Details:} Tiny Imagenet has 200 classes. Each class has 500 training images, 50 validation images, and 50 test images. Each image has a resolution of 64 pixels by 64 pixels by 3 channels.

\textbf{Changes to model parameters:} We utilize a similar model for ResNet18 and VGG19 as in the earlier experiment with CIFAR-10. We only modify the models slightly in order to cater for the 200 output classes of Tiny ImageNet, which is an increase from the 10 output classes in CIFAR-10. As such, the last linear layer for ResNet18 has 200 nodes (instead of 10 nodes for CIFAR-10), while the last two linear layers for VGG19 has 1024 nodes and 200 nodes respectively (instead of 256 nodes and 10 nodes respectively for CIFAR-10). 

Due to the complexity of this dataset, in order to attain better train/test accuracies, we apply image augmentation to each data sample per epoch. The image augmentation parameters applied are shown in Table \ref{table:dataaugment}.

\begin{table}[t]
	\caption{Image Augmentation Parameters}
	\label{table:dataaugment}
	\vskip 0.15in
	\begin{center}
		\begin{small}
			\begin{sc}
				\begin{tabular}{cc}
					\toprule
					Metric & Value\\
					\midrule
					Rotation & 40 degree\\
					Width shift range & 0.2\\
					Height shift range & 0.2\\
					Zoom range & 0.2\\
					Shear range & 0.2\\
					Flipping & Horizontal\\
					\bottomrule
				\end{tabular}
			\end{sc}
		\end{small}
	\end{center}
	\vskip -0.1in
\end{table}

In order to give the model more time to converge for this larger dataset, we also increase the number of epochs before early stopping to 10 (instead of 5 in earlier experiments).

Also, in order to reduce experimental running time for Tiny ImageNet, we drop at each pruning cycle a fraction 0.5 of the filters (instead of 0.2 for CIFAR-10).

\textbf{Experiments:} We conduct an experiment using Algorithm 1 for ResNet18 and VGG19 on Tiny ImageNet:

\begin{enumerate}[leftmargin=*,nosep, label=4.\arabic*)]
\item \textbf{ResNet18}. The plot of training and test accuracy against fraction of filters remaining for various metrics is shown in Figs. \ref{fig:train_resnet_tinyimagenet} and \ref{fig:test_resnet_tinyimagenet} respectively.
\item \textbf{VGG19}. The plot of training and test accuracy against fraction of filters remaining for various metrics is shown in Figs. \ref{fig:train_vgg_tinyimagenet} and \ref{fig:test_vgg_tinyimagenet} respectively.
\end{enumerate}

For ResNet18, it can be seen (Figs. \ref{fig:train_resnet_tinyimagenet} and \ref{fig:test_resnet_tinyimagenet})  that \texttt{minimum} metric and \texttt{minimum\_layer} metrics have the most competitive performance, followed by \texttt{random\_layer}, \texttt{random},  \texttt{maximum\_layer} and  and lastly \texttt{maximum} metric. The \texttt{minimum} metric is the most competitive. 

For VGG19, it can be seen (Figs. \ref{fig:train_vgg_tinyimagenet} and \ref{fig:test_vgg_tinyimagenet}) that the \texttt{minimum\_layer} metric performs the best, followed by \texttt{random}, \texttt{random\_layer}, \texttt{max\_layer},  \texttt{aximum},  and  and lastly \texttt{minimum} metric. The \texttt{minimum\_layer} metric is the most competitive. 

For both ResNet18 and VGG19,  \texttt{maximum} can be seen to be consistently poor when the fraction of filters remaining is 0.3 and below. Surprisingly, \texttt{minimum} has the poorest performance for VGG19.

\textbf{Evaluation:} The results show that for ResNet18, \texttt{minimum} performs the best. This is similar to the results obtained on the CIFAR-10 dataset (Refer to Figs. \ref{fig:train_resnet} and \ref{fig:test_resnet}). This could be the fact that the skip connections in ResNet18 allow the pruned model to perform well even if majority of the layer is removed, and gives greater redundancy for pruning. With such redundancy, some of the pitfalls of global pruning methods can be alleviated.

For VGG19, \texttt{minimum\_layer} performs the best, while \texttt{minimum} performs the worst. This could again be because there is no redundancy in the layers for VGG19 and if one layer gets pruned aggressively by a global metric, it might affect performance negatively. The drop in performance of \texttt{minimum} is worse in the Tiny ImageNet dataset as compared to the CIFAR-10 dataset (Refer to Figs. \ref{fig:train_vgg} and \ref{fig:test_vgg}) likely because the proportion of filters dropped per pruning cycle is larger at 0.5 as compared to 0.2, hence there is a greater chance of a layer getting pruned aggressively.

Similar to the CIFAR-10 experiments, the experiments on the larger Tiny ImageNet dataset also suggest that the \texttt{minimum} and \texttt{minimum\_layer} are both competitive in ResNet18. For most models, the layer-wise \texttt{minimum\_layer} metric is a general all-round metric to be used.

Using \textit{DropNet}, we can reduce the number of filters by 50\% or more without significantly affecting model accuracy, highlighting its effectiveness in reducing network complexity. 

\begin{figure*}[t]
\centering
	\begin{minipage}[t]{0.45\textwidth}
        \includegraphics[width=\textwidth]{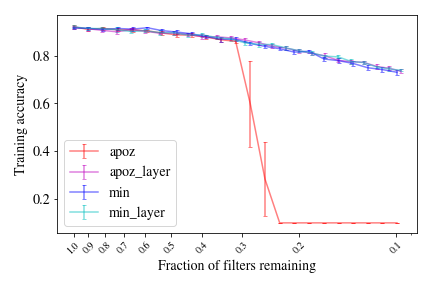}
		\caption{Plot of training accuracy against fraction of filters remaining for DropNet and APoZ in ResNet18 on CIFAR-10}
		\label{fig:train_resnet_apoz}
	\end{minipage}%
	\hfill
	\begin{minipage}[t]{0.45\textwidth}
		\includegraphics[width=\textwidth]{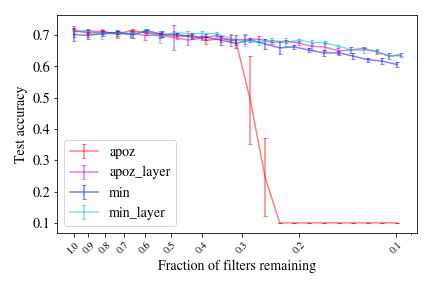}
		\caption{Plot of test accuracy against fraction of filters remaining for DropNet and APoZ in ResNet18 on CIFAR-10}
		\label{fig:test_resnet_apoz}
	\end{minipage}%
\end{figure*}

\begin{figure*}[t]
\centering
	\begin{minipage}[t]{0.45\textwidth}
        \includegraphics[width=\textwidth]{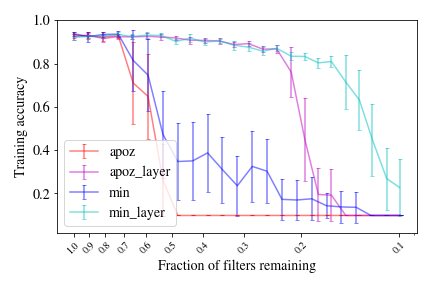}
		\caption{Plot of training accuracy against fraction of filters remaining for DropNet and APoZ in VGG19 on CIFAR-10}
		\label{fig:train_vgg_apoz}
	\end{minipage}%
	\hfill
	\begin{minipage}[t]{0.45\textwidth}
		\includegraphics[width=\textwidth]{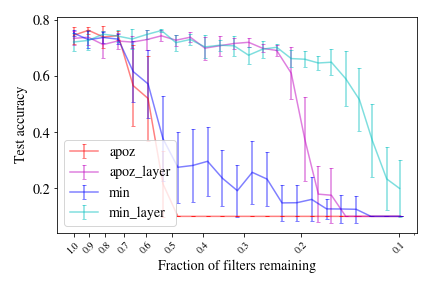}
		\caption{Plot of test accuracy against fraction of filters remaining for DropNet and APoZ in VGG19 on CIFAR-10}
		\label{fig:test_vgg_apoz}
	\end{minipage}%
\end{figure*}

\section{Benchmarking against APoZ}

\textit{\textbf{Q5. Does DropNet perform better than prior data-driven pruning methods?}}

While we utilize an oracle for smaller model sizes such as Model A and B, in order to evaluate the effectiveness of \textit{DropNet} for larger models, we compare its performance to a similar data-driven metric known as Average Percentage of Zeros (APoZ) \cite{hu2016data}.

APoZ measures the percentage of zero activations of a neuron after a ReLU activation function. The neuron/filter with the highest percentage of zero activations is considered least important and is pruned first.

In their original paper \cite{hu2016data}, APoZ used a variant of layer-wise pruning, where they first prune ``a few layers with high mean APoZ, and then progressively trim its neighboring layers". In order to keep the methodology consistent to that of \textit{DropNet}, we adapt the same APoZ metric of percentage of zero activations of a neuron/filter after a ReLU activation, but use \textit{DropNet}'s layer-wise and global-wise iterative pruning approaches as depicted in Algorithm 1. APoZ using layer-wise pruning is termed \texttt{apoz\_layer}, while APoZ using global pruning is termed \texttt{apoz}. We compare its performance to DropNet's layer-wise pruning \texttt{minimum\_layer} and global pruning \texttt{minimum}.

We conduct an experiment to compare \textit{DropNet} and APoZ using Algorithm 1 for ResNet18 and VGG19 on CIFAR-10:

\begin{enumerate}[leftmargin=*,nosep, label=5.\arabic*)]
\item \textbf{ResNet18}. The plot of training and test accuracy against fraction of filters remaining for \textit{DropNet} and APoZ is shown in Figs. \ref{fig:train_resnet_apoz} and \ref{fig:test_resnet_apoz} respectively.
\item \textbf{VGG19}. The plot of training and test accuracy against fraction of filters remaining for \textit{DropNet} and APoZ is shown in Figs. \ref{fig:train_vgg_apoz} and \ref{fig:test_vgg_apoz} respectively.
\end{enumerate}

For ResNet18, it can be seen (Figs. \ref{fig:train_resnet_apoz} and \ref{fig:test_resnet_apoz}) that the \texttt{minimum\_layer} and \texttt{apoz\_layer} both perform the best, followed closely by \texttt{minimum}, then the \texttt{apoz}. After a fraction of 0.7 or more filters are pruned, the \texttt{apoz} metric suffers a huge performance drop.

For VGG19, it can be seen (Figs. \ref{fig:train_vgg_apoz} and \ref{fig:test_vgg_apoz}) that the \texttt{minimum\_layer} performs the best, followed by \texttt{apoz\_layer}, then \texttt{minimum} and finally \texttt{apoz}. After a fraction of 0.3 or more filters are pruned, the global metrics \texttt{apoz} and \texttt{minimum} suffer a huge performance drop.

\textbf{Evaluation:} The results show that DropNet in general outperforms APoZ, both layer-wise and globally. In general, for the same amount of filters pruned, \textit{DropNet} achieves higher test accuracy than APoZ. The \texttt{minimum\_layer} metric is consistently the best performing metric across both ResNet18 and VGG19 models, outperforming \texttt{apoz\_layer}. For global metrics, the \texttt{minimum} metric also consistently outperforms \texttt{apoz}. Notably, while the \texttt{minimum} metric has good performance for ResNet18, the same performance is not seen in \texttt{apoz}.

This shows that \textit{DropNet} has merit as a data-driven pruning approach, as it captures more information about the importance of a particular node/filter through the use of the expected absolute value. This is an improvement from APoZ, as it also takes into account the magnitudes of the post-activation values, rather than just only relying on the percentage of zero activations of a node/filter.

\section{Concluding Remarks}

The results show that \textit{DropNet} shows significantly better performance than random pruning, even for larger models such as ResNet18 and VGG19. \textit{DropNet} also manages to achieve higher test accuracy for the same amount of pruning as compared to prior work APoZ, highlighting its competency. \textit{DropNet} is a highly-effective general-purpose pruning algorithm able to work on datasets of varying sizes such as MNIST, CIFAR-10 and Tiny ImageNet. Overall, \textit{DropNet} is able to prune up to 90\% or more of nodes/filters without significant loss of accuracy.

\textbf{Global or layer-wise pruning:} As shown in the main paper, if we are pruning small models such as Model A or Model B, \texttt{minimum} works well. Furthermore, we show here that in large models such as Model C, \texttt{minimum} shows promise in avoiding pruning bottlenecks as compared to its layer-wise counterpart \texttt{minimum\_layer}, and gives significantly better performance if we are just pruning a small fraction of the original model. However, when pruning even larger models such as ResNet18 and VGG19, we show that it is better to use \texttt{minimum\_layer} instead. One reason for this may be that the statistical properties of the post-activation values of each layer may differ significantly as the model grows large, and a global metric for all the layers may not work as well. That said, the empirical results of ResNet18 show that \texttt{minimum} can be competitive as well for these larger models, which suggests that skip connections may be able to alleviate the pitfalls of global metrics.

\end{document}